\documentclass[10pt,twocolumn,letterpaper]{article}

\usepackage[pagenumbers]{cvpr} 
\usepackage[accsupp]{axessibility}  

\definecolor{cvprblue}{rgb}{0.21,0.49,0.74}
\definecolor{mygrey}{RGB}{220,220,220}
\definecolor{mygrey2}{RGB}{235,235,235}
\usepackage[pagebackref,breaklinks,colorlinks,allcolors=cvprblue]{hyperref}
\usepackage{float}
\usepackage{multirow}
\usepackage{colortbl}

\def\paperID{9418} 
\def\confName{CVPR}
\def\confYear{2025}

\newcommand\ourbench{VinaBench}
\title{\ourbench{}: Benchmark for Faithful and Consistent Visual Narratives}

\author{
\textbf{Silin Gao$^{1}$, Sheryl Mathew$^{1,3}$, Li Mi$^{1}$, Sepideh Mamooler$^{1}$, Mengjie Zhao$^{2}$,} \\
\textbf{Hiromi Wakaki$^{2}$, Yuki Mitsufuji$^{2}$, Syrielle Montariol$^{1}$, Antoine Bosselut$^{1}$} \\
$^1$EPFL, Switzerland $^2$Sony Group Corporation, Japan $^3$Carnegie Mellon University, USA 
}

\begin{document}
\twocolumn[{%
\renewcommand\twocolumn[1][]{#1}%
\maketitle
\begin{center}
    \captionsetup{type=figure}
    \includegraphics[width=.99\textwidth]{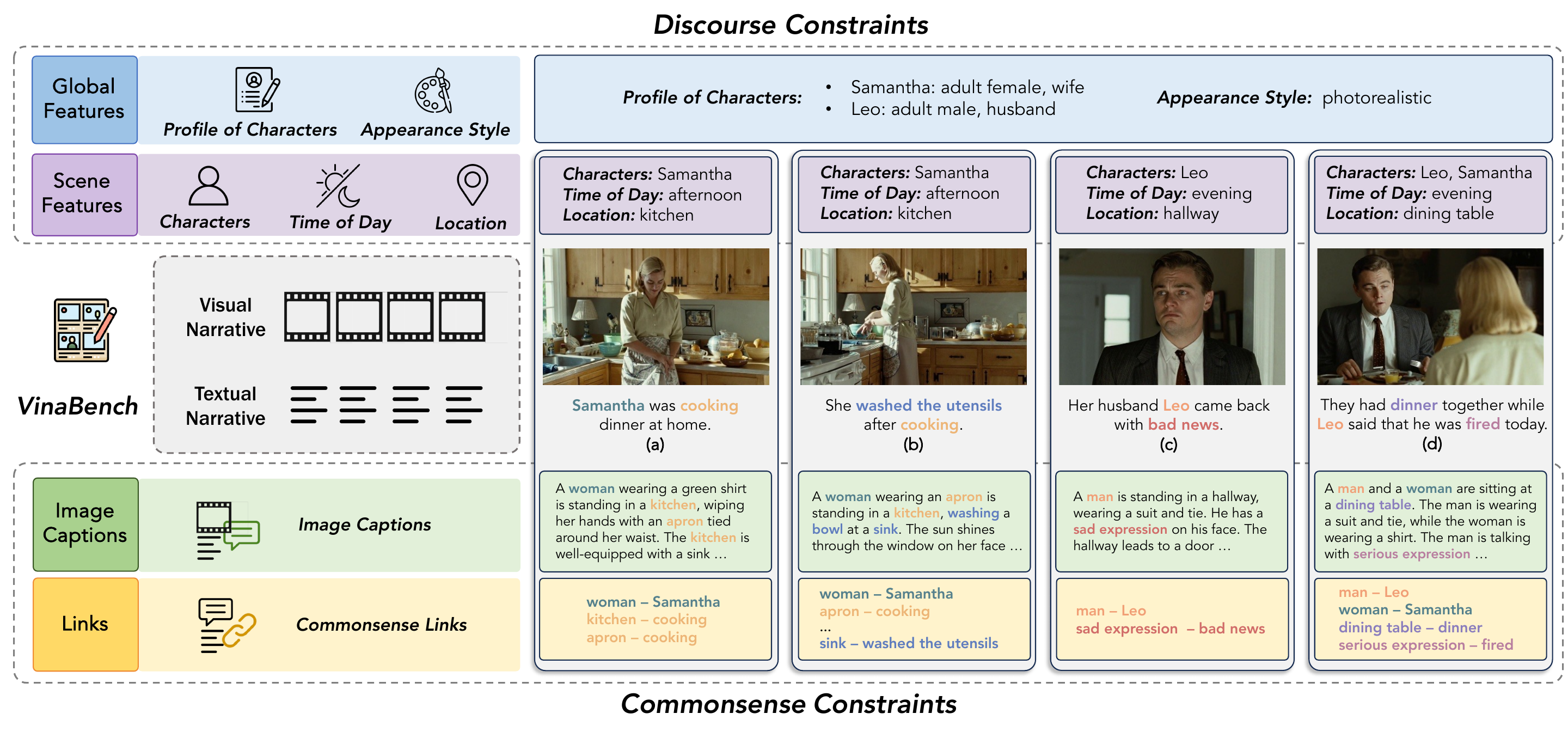}
    \captionof{figure}{\textbf{Overview of \ourbench{}.} We augment existing visual-textual narrative pairs with \textit{discourse and commonsense constraints}, to promote the learning of consistent and faithful visual narrative generation and its evaluation. The \textit{commonsense constraints} consist of links that ground the visual entities (extracted from image captions) to their associated textual narrative entities, as labeled by the phrases paired with the same color. The \textit{discourse constraints} include scene-specific narrative features that trace the dynamics of basic narrative elements, \ie{}, characters, time and location, and global narrative features that describe static character attributes and image appearance style.}
    \label{data_overview}
\end{center}
}]


\begin{abstract}

Visual narrative generation transforms textual narratives into sequences of images illustrating the content of the text.
However, generating visual narratives that are faithful to the input text and self-consistent across generated images remains an open challenge, due to the lack of knowledge constraints used for planning the stories.
In this work, we propose a new benchmark, \ourbench{}, to address this challenge.
Our benchmark annotates the underlying commonsense and discourse constraints in visual narrative samples, offering systematic scaffolds for learning the implicit strategies of visual storytelling.
Based on the incorporated narrative constraints, we further propose 
novel metrics to closely evaluate the consistency of generated narrative images and the alignment of generations with the input textual narrative.
Our results across three generative vision models demonstrate that learning with \ourbench{}'s knowledge constraints effectively improves the faithfulness and cohesion of generated visual narratives.\footnote{We release our data and code to the community, our project page: \url{https://silin159.github.io/Vina-Bench}}

\end{abstract}    
\section{Introduction}

\label{sec:intro}

Human narratives are often transformed from text into visual media, \eg{}, in the film and television industries, scripts written by screenwriters are usually visualized as storyboards by art designers, to assist the filming of movies and TV series.
%
However, translating textual narratives to sequences of images requires addressing two fundamental challenges: \textit{narrative alignment} and \textit{visual consistency}.

First, as textual narratives are often abstract and visually under-specified, visual narrative generation models must infer relevant commonsense knowledge to manifest relevant and coherent visual content.
For example, in Frame (c) of Figure~\ref{data_overview}, the phrase \textit{bad news} in the textual narrative is visually interpreted as a \textit{sad expression} on the husband Leo's face.
This visual interpretation of the character's state of mind is not explicitly mentioned in the textual narrative, demonstrating the \textbf{manifestation gap} between the input textual narrative and output visual narrative.
Second, visual narratives possess discourse features \citep{cohn2013visual,cohn2017drawing}, \ie{}, narrative elements such as characters and locations, that may be connected across different images of the visual narrative.
For instance, Figure~\ref{data_overview}~(a) and (b) are closely connected in the visual discourse, where the basic settings of the scene remain the same, \ie{}, Samantha staying in a kitchen.
Visual narrative generation models must plan such visual discourse, and be consistent in how these various features are manifested.

However, previous methods typically do not explicitly address these challenges for visual narrative generation \citep{maharana2022storydall,pan2024synthesizing,zheng2024contextualstory,liu2024intelligent}, and instead simply learn to map text narratives directly to visual narratives.
Consequently, they do not model the necessary commonsense knowledge for producing visual manifestations from the narrative context, and are therefore prone to generate images that are not faithful to the narrative.
They also fall short of learning the consistency constraints in visual narrative discourse, often generating image sequences with inconsistent character appearances, background location, or time period.\footnote{as verified by our analysis in \S\ref{sec:results}}

In this work, we propose a benchmark to address the aforementioned challenges in visual narrative generation, which augments visual narrative exemplars with commonsense and discourse constraints, as illustrated in Figure~\ref{data_overview}.
Our \textbf{Vi}sual \textbf{na}rrative \textbf{Bench}mark, \textbf{\ourbench{}}, contains $\sim$25K pairs of visual and textual narratives sampled from diverse visual storytelling datasets \citep{hong2023visual,xie2024learning,liu2024intelligent}.
\ourbench{} also contains commonsense links that bridge the manifestation gap between textual and visual narratives, which enables better learning of their commonsense alignment.
Specifically, the fine-grained content in visual narratives is first extracted as image captions, whose entities (noun or verb phrases) are then linked to their associated textual narrative entities.
Moreover, \ourbench{} annotates a set of global and scene-specific features to explicitly reveal the visual discourse.
The global features describe the static attributes of characters and the image appearance style.
The scene (per image) features trace the dynamics of basic narrative elements, including presented characters, time of day and location.
These discourse features promote visual narrative consistency, and the alignment of scene dynamics to narrative progression.

Our benchmark evaluation uses commonly-adopted metrics for matching generated visual narrative images to gold references, based on Frechet inception distance \citep{heusel2017gans}, or CLIP \citep{radford2021learning} similarity score of the two modalities, etc.
However, these metrics may be biased to specific reference images, which are not the only feasible visual manifestations of their corresponding narrative, \eg{}, the woman in Figure~\ref{data_overview}~(a) was not necessarily wearing a green shirt.
To address this limitation, we also propose a novel set of evaluation metrics for visual narrative generation that highlights the consistency and manifestation assessment of key narrative elements, labeled by our constructed visual discourse and commonsense constraints.
Our proposed metrics are either reference-free or based on ranking a pool of sampled image candidates, mitigating the impact of single reference comparisons that might skew the evaluation to irrelevant details.

Using these new resources, we test several representative visual narrative generation models \citep{tian2024mm,pan2024synthesizing,liu2024intelligent} on \ourbench{}.
Our results on all models consistently show that learning with our constructed discourse and commonsense constraints significantly augments the visual narrative consistency and alignment to the input textual narrative.
However, all of our tested models still have large room for improvement when comparing to human-crafted references, which calls for further research on developing better visual narrative generation methods.

\section{Background and Related Work}
\label{sec:related}

\paragraph{Visual Narrative Generation}
Transforming textual narratives into image sequences requires manifesting visual elements that are implied, though rarely explicitly stated, over the course of storytelling, which is essential for understanding and generating longer videos \citep{li2019storygan,xie2024learning}.
More intuitive visual illustrations also benefit the education of complex real-world concepts, and contribute to the childhood development of intelligence, imagination and creativity \citep{dickinson2012reading,strouse2018role}.

Current visual narrative generation methods \citep{maharana2022storydall,pan2024synthesizing,zheng2024contextualstory,liu2024intelligent} mostly rely on pre-trained vision transformers \citep{ramesh2021zero,radford2021learning,li2022blip} and diffusion modules \citep{ramesh2022hierarchical,rombach2022high,saharia2022photorealistic} to model direct textual-to-visual narrative mapping, which often fall short of learning the underlying commonsense and discourse constraints of this task.
Although prior works \citep{maharana2021integrating,chen2022character,li2022learning} have stepped into the commonsense augmentation and alignment in visual story generation, they are limited to simple physical commonsense in ConceptNet \citep{speer2017conceptnet} and general word or token-level semantic alignment with image regions, which overlooks more in-depth commonsense alignment between textual and visual expression manners.

Besides, the image sequences commonly studied in visual narrative generation are formed by either photos from different origins \citep{huang2016visual}, or video shots from a single cartoon \cite{li2019storygan,maharana2021integrating}, which only cover pseudo or monotonous visual narrative cases.
As more real and diverse visual narrative data resources \citep{hong2023visual,xie2024learning,liu2024intelligent} recently emerge, our work aims to annotate the commonsense and discourse constraints implied in these visual narrative resources, and provide benchmark methods of augmenting visual narrative generation with our incorporated constraints.

\paragraph{Visual-Linguistic Alignment}
Linking visual data with its natural language correspondence contributes to robust modeling of world visual concepts \citep{radford2021learning}, which promotes the advancement of various vision-language applications, \eg{}, visual question answering \citep{antol2015vqa}, visual dialogue \citep{das2017visual}, and visual storytelling \citep{huang2016visual}.
Due to the need for more refined vision understanding in the above applications, fine-grained visual-linguistic alignment techniques are studied, \eg{}, matching visual scene graphs with linguistic structures \citep{nie2021triangle,wang2020storytelling}, aligning image patches with text tokens or physical knowledge graphs \citep{li2022fine,maharana2021integrating,xiong2022mga}, etc.
Different from prior works, we focus on more implicit commonsense alignment between visual expressions and textual descriptions, in the context of visual narrative generation.

\paragraph{Visual Narrative Structure}
Natural language possesses syntactic structures \citep{lees1957syntactic}, \ie{}, words in a sentence have their lexical categories, \eg{}, Noun (N), Verb (V), etc., and can be further grouped into higher-level phrases such as Noun Phrase (NP), Verb Phrase (VP), etc.
Similarly, visual narratives also possess structures \citep{cohn2013visual}, where images in a visual narrative can be mapped into five categories according to the tension of the narrative, including Establisher (E), Initial (I), Prolongation (L), Peak (P) and Release (R).
These categories then form phases of constituency, \eg{}, a canonical constituency phase consists of a linear order of the categories E-I-L-P-R.
Based on the above structure, a visual narrative can be divided into discourse segments \citep{cohn2017drawing}, whose boundaries are determined by the start and end of the narrative's constituency phases.
The discourse segments feature the dynamics of narrative elements across different images (or scenes), \ie{}, typically the persistence and change of characters, time and spatial location, which are the key information intuitively perceived by narrative viewers \citep{magliano2001indexing,zacks2009segmentation,magliano2011impact}.
In this work, we aim to concretize the discourse dynamics of visual narratives, and study how they enhance the consistency of visual narrative generation.

\paragraph{Visual Consistency Evaluation}
Related to our focus on visual narrative consistency, research on video generation \citep{huang2024vbench,liao2024evaluation} raises the evaluation of semantics and style consistency, \wrt{} attributes and spatial relationships of objects, actions of characters, temporal and appearance style, etc.
Different from video consistency which focuses more on the short-time spatial-temporal coherence, our work is more concerned with the long-time visual element consistency throughout the narrative discourse.

\begin{figure}[t]
\centering
\includegraphics[width=1.0\columnwidth]{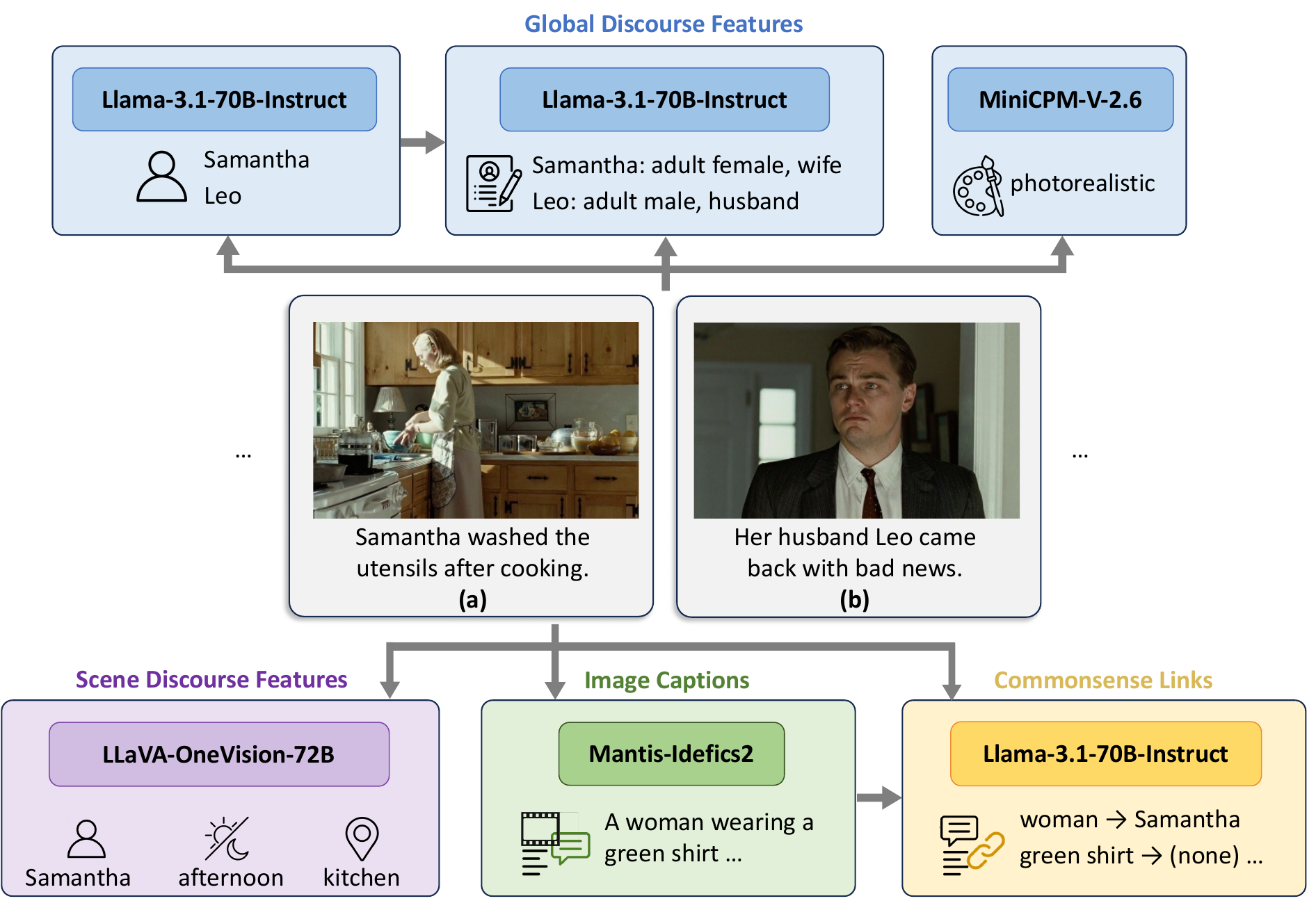}
\caption{\textbf{Overview of \ourbench{} data construction pipeline.} We use hybrid VLMs and LLMs to annotate the discourse features and commonsense links underlying visual-textual narrative pairs.}
\label{data_construction_fig}
\end{figure}

\section{\ourbench{} Data Construction}
\label{sec:data}

\ourbench{} samples visual-textual narrative pairs from three advanced visual storytelling datasets, Visual Writing Prompts (VWP) \citep{hong2023visual}, Storyboard20K \citep{xie2024learning} and StorySalon \citep{liu2024intelligent}, which cover diverse characters and scenes.
Using these datasets as a foundation, as illustrated in Figure~\ref{data_overview}, we augment the visual-textual narrative pairs with commonsense and discourse constraints.
These constraints highlight the alignment between visual and textual narrative manifestations, and the consistency of basic elements expressed in the visual narrative.
Figure~\ref{data_construction_fig} summarizes our components for constructing commonsense and discourse constraints given a visual narrative.

\subsection{Commonsense Constraints}
\label{subsec:commonsense}
Commonsense constraints are entity links that ground the visual details in narrative images to their relevant textual narrative phrases, \eg{}, the woman in Figure~\ref{data_construction_fig}~(a) is linked to the character Samantha.
We extract the commonsense entity links in three steps:

First, we use dense captioning \citep{johnson2016densecap} to extract visual details in each narrative image.
We prompt Mantis-Idefics2 \citep{jiang2024mantis} to generate the dense captions, which achieves outstanding performance among various VLMs \cite{zhu2023minigpt,peng2023kosmos,ma2025groma,yao2024minicpm} in our pilot study.
We input each narrative image with its textual narrative description as context, which effectively prevents the model from generating hallucinated details in the dense caption that contradict the textual narrative.
For instance, in Figure~\ref{data_construction_fig}~(a), without knowing the textual narrative, the model may conclude that the woman is stirring food in the bowl, instead of washing the bowl.

Second, we extract visual entities in the generated dense caption of each narrative image.
We prompt Llama3.1-70B-Instruct (Llama3.1) \citep{dubey2024llama}, a powerful open source LLM, to perform the extraction, where the target visual entities are scoped to noun or verb phrases presented in the caption.

Finally, for each visual entity extracted from the image caption, we find its potential commonsense link to the entities (noun or verb phrases) in the textual narrative.
In particular, we present Llama3.1 with the textual narrative and the visual entity contextualized by its source image caption, and prompt the model to find an entity from the narrative that is associated with the visual entity.
For visual entities that do not link to any entity in the textual narrative, \eg{}, \textit{green shirt} in the image caption of Figure~\ref{data_construction_fig}~(a), we instruct Llama3.1 to output an empty link \textit{``no link''}.

\subsection{Discourse Constraints}
\label{subsec:discourse}
We parse a group of global and scene-specific features to represent discourse constraints for each narrative.
Our annotated features identify discourse concepts from previous studies of visual narrative structure \citep{cohn2013visual,cohn2017drawing}, and also considerations for style consistency \citep{huang2024vbench}, as described in Section~\ref{sec:related}.
Below, we introduce the frame and construction of our discourse features.

\paragraph{Frame}
We annotated two varieties of global features:

\begin{itemize}
\item \textbf{Character Profiles} includes the full list of characters involved in the narrative.
Each character is indexed by his or her name (if the name is not mentioned, a role pronoun such as \textit{man} or \textit{woman} is used instead), and described by basic attributes including age range (\eg{}, \textit{young adult}, \textit{child}), gender (\textit{male} or \textit{female}), social role (\eg{}, \textit{husband}, \textit{Tom's close friend}), and other sustained physical features (\eg{}, \textit{badly hurt}).
Basic character attributes are expected to remain static over the course of narrative.
\item \textbf{Appearance Style} describes the style \citep{huang2024vbench} of the visual narrative, \eg, \textit{photorealistic}, \textit{fantasy art}, \textit{digital art}, \textit{pop art}, \textit{comic book}, \textit{cartoon}, \textit{surrealistic} and \textit{black and white photographic}.
Most visual narratives typically maintain a consistent appearance style across narrative images.
If the images of a visual narrative are found to have multiple styles, this label will be annotated as \textit{not unified}.
\end{itemize}

\noindent The scene-specific features of each image in the visual narrative consist of three components:

\begin{itemize}
\item \textbf{Characters} that are presented in the image, whose appearances are expected to align with their descriptions in the global profile.
A character's appearance typically remains consistent across images where they are presented.
\item \textbf{Time of Day} indicates the period of day during which the scene occurs, including \textit{early morning}, \textit{morning}, \textit{afternoon}, \textit{evening} and \textit{night}, which may shift as the narrative progresses.
The time of day is labeled as \textit{unclear} if it is ambiguous in the scene (\eg, if the scene is indoors).
\item \textbf{Location} describes where the scene takes place, \eg{}, \textit{kitchen}, \textit{restaurant}, \textit{outdoor road}, etc., or \textit{unclear} if ambiguous, which may also change dynamically during the narrative.
Images that are labeled with the same location are expected to have consistent spatial background.
\end{itemize}

\paragraph{Construction of Discourse Features}
We construct the global character profile in two steps.
We first prompt Llama3.1 to identify all characters in the narrative.
We input the entire textual narrative, and instruct the model to output a list of character names (or role pronouns) involved in the narrative.
Based on the identified character list, we then prompt the Llama3.1 to parse the basic attributes of each character in the list, given the entire textual narrative as context.
Note that we do not include the visual narrative or its captions in the context, to prevent the model from generating visual details that are not static or necessary attributes of the character, \eg{}, the woman \textit{Samantha} in Figure~\ref{data_overview} (a) has golden curly hair.

Based on the global character profile, we then label the presented characters in each scene, using a fine-grained two-step prompting strategy.
For each image in the visual narrative, we first prompt LLaVA-OneVision-72B (LLaVA-OV) \citep{li2024llava}, an advanced VLM with robust fine-grained multi-modal reasoning performance, to detect the number of characters in the image.
With the character number, the model is then instructed to further specify the detected characters' indexes (names or role pronouns) in the global profile, by matching their attributes to the content of the image and its corresponding textual description.
We also input the previous textual narrative as context, to resolve the issue of co-reference, \eg{}, \textit{``Her''} in Figure~\ref{data_construction_fig}~(a) refers to \textit{Samantha's}.

For the rest of discourse feature labels, we simply prompt LLaVA-OV to annotate the time of day and the location of each image in the narrative, given the image's corresponding textual description as context.
And we prompt MiniCPM-V-2.6 \citep{yao2024minicpm}, an advanced VLM optimized for multi-image understanding, to judge the image appearance style of the entire visual narrative.



\subsection{Expert Study}
\label{sec:data_analysis}


One question that naturally arises is whether the LLMs and VLMs used constructing \ourbench{} accurately annotated the constraints of visual narrative samples.
To evaluate this, 12 experts manually check the labels of commonsense and discourse constraints of 100 narrative samples in \ourbench{}.
For each narrative sample, the experts first check whether its global discourse features appropriately depict the attributes of characters in the narrative and the appearance style of the narrative images.
Then, for a specific scene randomly selected from the narrative sample, the experts check whether its (scene-specific) features correctly label its presented characters, time of day and location, and whether its image caption and commonsense links reasonably describe its image content and associations to its textual narrative description, respectively.
Each narrative sample is checked by two experts, and we report their average rate of accepting the labels as correct or appropriate, with the percentage of their disagreements.

Table~\ref{tab:expert_study} shows the results of our expert study.
We observe high acceptance rates for all types of constraint labels, each with fairly low rates of disagreement between the experts.
These results verify that \ourbench{} construction scheme using large language and vision models is reliable for annotating accurate visual narrative constraints, which saves the labour of human annotators.

\begin{table}[t]
\centering
\resizebox{1.0\columnwidth}{!}{
\smallskip\begin{tabular}{lccccccc}
\toprule
\textbf{Rate (\%)} & \textbf{Sty.}  & \textbf{Attr.} & \textbf{Cap.} & \textbf{CL} & \textbf{Pre.}  & \textbf{Time} & \textbf{Loc.}\\
\midrule
\textbf{Accept}   & 95.5 & 91.1 & 85.0 & 86.6 & 84.5 & 89.0 & 93.0 \\
\textbf{Disagree} & 3.0 & 5.0 & 8.0 & 6.0 & 6.0 & 8.0 & 4.0 \\
\bottomrule
\end{tabular}
}
\caption{Expert study on the accuracy of commonsense and discourse constraints labeled in \ourbench{}, including appearance style (\textbf{Sty.}), character attributes (\textbf{Attr.}), image caption (\textbf{Cap.}), commonsense links (\textbf{CL}), presented characters (\textbf{Pre.}), time of day (\textbf{Time}) and location (\textbf{Loc.}). Experts' average acceptance rate (\textbf{Accept}) and percentage of disagreement (\textbf{Disagree}) are reported.}
\label{tab:expert_study}
\end{table}

\section{\ourbench{} Evaluation}
\label{sec:metrics}

Prior work in visual narrative generation \citep{maharana2022storydall,pan2024synthesizing,zheng2024contextualstory,liu2024intelligent} evaluated models on full-reference metrics, \eg{}, FID \citep{heusel2017gans}, which directly match model generations to gold reference images.
However, the visual expression of a narrative is always open-ended, \ie{}, not limited to a single reference.
Therefore, model generations that do not match references may receive lower scores, but still be acceptable manifestations of the textual narrative. For example, in Figure~\ref{data_overview}~(a), the model could visualize the woman with black hair instead of golden hair and remain faithful to the textual narrative. 
StoryGen \citep{liu2024intelligent} moved beyond reference images by checking CLIP \citep{radford2021learning} text-image similarity (CLIP-T) between model generations and the input textual narrative.
However, the mapping from CLIP similarity to the level (or rank) of alignment may vary across various narrative samples, \eg{}, if the input text is concise or under-specified, a vague similarity with CLIP-T $0.6$ may already indicate an outstanding level of alignment, while for relatively detailed input text, a high similarity with CLIP-T $0.9$ may be the outstanding bar instead.
%
Importantly, neither of the above metrics evaluates visual narrative consistency. Instead, they individually evaluate each generated image in the visual narrative, ignoring the inter-connections between different images, which remains assessed solely through human evaluation \citep{liu2024intelligent}.

Motivated by these shortcomings, we propose novel evaluation metrics to assess \textbf{visual-textual narrative alignment} and \textbf{visual narrative consistency}.
In particular, we design a \textit{ranking-based} metric (instead of fixed-range scoring) to measure more intuitive and uniform level of alignment between visual narrative generations and textual narrative inputs.
Based on our constructed commonsense and discourse constraints in Sec.~\ref{sec:data}, we further build a series of VQA-based \citep{lin2024evaluating} metrics to assess the \textit{fine-grained alignment} between visual generations and narrative constraints, and also the \textit{consistency} of visual narrative generations.
All of our metrics prevent the biases of directly comparing to a single reference image.
We describe our metrics below.

\subsection{Alignment Ranking}
\label{subsec:metrics_ranking}
We define a function $f(\cdot,\cdot)\in[0,1]$ to measure the pairwise alignment between an image and a textual description.
We test two implementations of the function, including CLIP \citep{radford2021learning} text-image embedding cosine similarity (CLIP-T), and VQAScore \citep{lin2024evaluating} where we ask LLaVA-OneVision-72B \citep{li2024llava} whether the image is aligned with the textual description (only answer \textit{Yes} or \textit{No}), and record the probability of the model outputting \textit{Yes} as its first decoded token.


For each scene, we use our defined function to sample top-$100$ images that have the highest alignment score with the input textual narrative, from the entire pool of images in the test set.
We then use the same function to score the alignment of the generated image with the input textual narrative, and use this score to obtain the generated image's ranking in the pool of sampled top-$100$ images.
For each model, we report the mean reciprocal rank (MRR) of its generated images across all scenes in the test set.
We denote our ranking metrics as \textbf{CLIP-T-MRR} and \textbf{VQA-MRR}, for CLIP-T and VQA-based ranking function, respectively.

\subsection{Fine-Grained Alignment}
\label{subsec:metrics_alignment}
We develop five metrics to measure the fine-grained alignment of each generated image with its corresponding scene's narrative constraints constructed in \ourbench{}.
\begin{itemize}
\item \textbf{Non-Character}: For each essential non-character entity in the scene's textual narrative, \ie{}, phrase that is linked in Sec. \ref{subsec:commonsense} but not included in the global character profile in Sec. \ref{subsec:discourse}, we prompt a VLM to judge whether the generated image contains or implies the phrase.
\item \textbf{Character Number}: We prompt a VLM to check whether the number of characters in the generated image matches the number of presented characters indicated by the scene's discourse feature.
\item \textbf{Character Attribute}: Given the scene's presented characters and their attributes in the global profile, we instruct a VLM to check whether characters depicted in the generated image fit into the given attributes.
\item \textbf{Time of Day}: If the time of day is not labeled as \textit{unclear} in the scene's discourse feature, we instruct a VLM to judge whether the image is taken during the labeled time.
\item \textbf{Location}: We instruct a VLM to judge whether the image is taken at the location labeled in the scene's discourse feature, if it is not \textit{unclear}.
\end{itemize}

\subsection{Consistency}
\label{subsec:metrics_consistency}
For each narrative sample, we design three metrics to assess the consistency of generated visual narrative images, based on our constructed features for the discourse constraints.

\begin{itemize}
\item \textbf{Style}: We prompt a VLM to judge whether all generated images in the narrative sample have the same appearance style. Note that the appearance style of generated images does not necessarily need to match the style labeled in the global discourse features, since the input textual narrative typically does not provide a constraint for image style.
\item \textbf{Character}: For each character in the global profile, if he or she is presented in multiple scenes according to the scene-specific discourse features, we instruct a VLM to check whether the generated images for those multiple scenes all show that same character.
\item \textbf{Location}: If multiple scenes possess the same location label in the scene-specific discourse features, we prompt a VLM to check whether the generated images for those multiple scenes are all taken at that same location.
\end{itemize}

\noindent For each fine-grained alignment and consistency metric, we follow VQAScore \citep{lin2024evaluating} to report the average probability of the VLM outputting \textit{Yes} as its first decoded token (the VLM is instructed to only answer \textit{Yes} or \textit{No}), under the zero-shot setting.
To ensure that our metrics are not biased on a specific VLM's preference, we run on two VLMs, MiniCPM-V-2.6 \citep{yao2024minicpm} and LLaVA-OneVision-72B \citep{li2024llava}, and confirm that the results given by the two VLMs are aligned.

\begin{table*}[t]
\centering
\resizebox{1.0\textwidth}{!}{
\smallskip\begin{tabular}{llcccccccccccc}
\toprule
\multirow{2}*{\textbf{Model}} & \multirow{2}*{\textbf{Setting}} & \multirow{2}*{\textbf{FID}}  & \multirow{2}*{\textbf{CLIP-I}} & \multirow{2}*{\textbf{CLIP-T}} & \multirow{2}*{\textbf{CLIP-T}} & \multicolumn{5}{c}{\textbf{Alignment}} & \multicolumn{3}{c}{\textbf{Consistency}} \\
    \cmidrule(lr){7-11}  \cmidrule(lr){12-14} 
 &  &  &  &  & \textbf{-MRR} & \textbf{Ent.} & \textbf{Num.} & \textbf{Attr.} & \textbf{Time} & \textbf{Loc.} & \textbf{Sty.} & \textbf{Char.} & \textbf{Loc.} \\
\toprule
\multirow{3}*{\textbf{ARLDM}} & \textbf{No Constraint} & 42.6 & 0.638 & 0.195 & 0.110 & 0.564 & 0.398 & 0.320 & 0.443 & 0.376 & 0.466 & 0.379 & 0.376 \\
                              & \textbf{LLM Constraints} & \textbf{37.6} & \textbf{0.676} & \textbf{0.204} & \textbf{0.151} & \textbf{0.674} & 0.443 & 0.411 & \textbf{0.512} & 0.584 & 0.859 & 0.551 & 0.689 \\
                              & \textbf{Gold Constraints} & \underline{35.3} & \underline{0.716} & \underline{0.209} & \underline{0.155} & \underline{0.682} & 0.619 & \underline{0.546} & \underline{0.518} & 0.690 & 0.854 & 0.569 & 0.697 \\
\midrule
\multirow{3}*{\textbf{StoryGen}} & \textbf{No Constraint} & 78.6 & 0.562 & 0.184 & 0.100 & 0.471 & 0.335 & 0.285 & 0.279 & 0.315 & 0.238 & 0.231 & 0.311 \\
                              & \textbf{LLM Constraints} & 52.1 & 0.600 & 0.190 & 0.106 & 0.595 & 0.424 & 0.341 & 0.367 & 0.504 & 0.452 & 0.418 & 0.465 \\
                              & \textbf{Gold Constraints} & 48.9 & 0.619 & 0.194 & 0.115 & 0.614 & 0.547 & 0.444 & 0.393 & 0.598 & 0.475 & 0.423 & 0.527 \\
\midrule
\multirow{8}*{\textbf{MM-Inter.}} & \textbf{No Constraint} & 48.3 & 0.634 & 0.176 & 0.066 & 0.499 & 0.409 & 0.326 & 0.463 & 0.449 & 0.947 & 0.582 & 0.449 \\
                              & \textbf{LLM Constraints} & 42.2 & 0.667 & 0.198 & 0.111 & 0.643 & \textbf{0.458} & \textbf{0.412} & 0.486 & \textbf{0.600} & \textbf{0.986} & \textbf{0.678} & \textbf{0.764} \\
                              & \quad - \textbf{w/o CL} & 42.9 & 0.666 & 0.197 & 0.109 & 0.642 & 0.455 & 0.409 & 0.475 & 0.597 & 0.983 & 0.643 & 0.758 \\
                              & \quad - \textbf{w/o DF} & 43.3 & 0.657 & 0.196 & 0.107 & 0.624 & 0.449 & 0.401 & 0.474 & 0.564 & 0.978 & 0.644 & 0.684 \\
                              & \quad - \textbf{w/o GDF} & 42.6 & 0.665 & 0.197 & 0.110 & 0.642 & 0.450 & 0.407 & 0.484 & 0.597 & 0.978 & 0.649 & 0.760 \\
                              & \quad - \textbf{w/o SDF} & 42.6 & 0.663 & 0.196 & 0.109 & 0.634 & 0.449 & 0.411 & 0.477 & 0.574 & 0.979 & 0.673 & 0.685 \\
                              & \quad - \textbf{Random} & 53.7 & 0.614 & 0.174 & 0.048 & 0.415 & 0.399 & 0.318 & 0.412 & 0.412 & 0.945 & 0.576 & 0.447 \\
                              & \textbf{Gold Constraints} & 39.3 & 0.698 & 0.200 & 0.118 & 0.652 & \underline{0.623} & \underline{0.546} & 0.497 & \underline{0.728} & \underline{0.976} & \underline{0.688} & \underline{0.856} \\
\midrule
\rowcolor{mygrey} \textbf{Gold Ref.}  & - & - & - & 0.208 & 0.159 & 0.776 & 0.813 & 0.758 & 0.756 & 0.863 & 0.971 & 0.780 & 0.863 \\
\bottomrule
\end{tabular}
}
\caption{Evaluation results on \textbf{VWP} narratives. The displayed results of our VQA-based metrics are rooted on MiniCPM-V-2.6, \wrt{} the \textbf{Alignment} of non-character entities (\textbf{Ent.}), character number (\textbf{Num.}), character attributes (\textbf{Attr.}), time of day (\textbf{Time}) and location (\textbf{Loc.}), and the \textbf{Consistency} of style (\textbf{Sty.}), character (\textbf{Char.}) and location (\textbf{Loc.}). \textit{Gold Ref.} denotes gold references. \colorbox{mygrey2}{Best results with LLM Constraints and with Gold Constraints are \textbf{bolded} and \underline{underlined}, respectively.} Lower FID score is better.}
\label{tab:results_main}
\end{table*}

\section{Experimental Methods}
\label{sec:baselines}

We evaluate various baseline visual narrative generation methods on \ourbench{}, based on a variety of task settings, models and metrics described below. We consider three settings to investigate augmenting the visual narrative generation model with the narrative constraints in \ourbench{}.
\begin{itemize}
\item \textbf{No Constraint}:
We first test a vanilla setting where the vision model is trained to generate the visual narrative images given only the textual narrative.
\item \textbf{LLM Constraints}:
We train a LLM, Llama3.1-70B-Instruct \citep{dubey2024llama} with LoRA \citep{hu2021lora}, to generate the constraints of each visual narrative scene, \ie{}, the scene's image caption and its corresponding commonsense links and discourse features constructed in Sec.~\ref{sec:data}, based on the textual narrative.
Then the vision model learns to generate the visual narrative images given the concatenation of textual narrative and LLM-generated constraints.
To enable training the auto-regressive LLM as a narrative constraint generator, we merge the commonsense links into the image caption, and concatenate it with the serialized discourse features, \eg{}, the narrative constraints of Figure~\ref{data_overview}(a) are serialized into \textit{A woman (Samantha) wearing a green shirt ... [Characters] Samantha (adult female, wife) [Time of Day] afternoon [Location] kitchen}.\footnote{Details of constraint preprocessing are in the supplementary material.}
\item \textbf{Gold Constraints}: We also test an oracle setting, where we replace the LLM-generated narrative constraints with our annotated gold constraints (with the same preprocessing of merging the commonsense links into the image caption and serializing the discourse features) at the inference phase.
\end{itemize}

\noindent We test three generative vision models that were optimized for visual narrative generation: ARLDM \citep{pan2024synthesizing}, StoryGen \citep{liu2024intelligent} and MM-Interleaved (MM-Inter.) \citep{tian2024mm}. 
We include detailed information about the three vision models in our supplementary material.
We evaluate model generations based on our proposed \textbf{Alignment} and \textbf{Consistency} metrics in Section~\ref{sec:metrics}, as well as previously reported metrics commonly used for visual narrative generation: Frechet inception distance (\textbf{FID}) \citep{heusel2017gans}, and CLIP \citep{radford2021learning} embedding similarity to the gold reference image (\textbf{CLIP-I}) and to the narrative text (\textbf{CLIP-T}).

\section{Experimental Results}
\label{sec:results}
We first train and test our baseline models on \ourbench{}'s VWP movie narratives, and then evaluate their zero-shot generalization to the Storyboard20K testing samples, which cover broader movie scenes and real movie synopses.
We also train and test all models on \ourbench{}'s StorySalon animation narratives, whose images have far different styles compared to the images from VWP and Storyboard20K, and are sourced from YouTube videos and E-books that are not limited to movies.

\subsection{VWP and Storyboard20K Narratives}
Table~\ref{tab:results_main} shows the evaluation results on the VWP narratives of \ourbench{}.\footnote{We include the results of our ranking-based metric VQA-MRR and fine-grained VQA-based metrics using LLaVA-OneVision-72B in the supplementary material, which indicate the same conclusions.}
We draw consistent conclusions on the three baseline models.
In particular, we find that inferring the narrative constraints before generating the narrative images (LLM Constraints) significantly improves the alignment and consistency of visual narrative generation.
This result suggests that the expressive gap between visual and textual narratives can be significantly bridged by middle-stage visual narrative planning, showing the importance of learning implicit visual narrative constraints, which can serve as an intermediate scaffold for visual narrative generation.

Interestingly, we find that the ranking scores (CLIP-T-MRR) of gold references (Gold Ref.) fall short of the maximum score (1.0), confirming that textual narratives typically do not map to single feasible visual narrative counterparts, and that full-reference metrics for visual narrative generation may be inadequate.
However, the model-generated visual narratives significantly lag the gold references on all metrics, indicating a large room of improvement. 

\begin{table}[t]
\centering
\resizebox{1.0\columnwidth}{!}{
\smallskip\begin{tabular}{@{~}l@{~~}l@{~~~}c@{~~}c@{~~}c@{~~}c@{~~}c@{~}}
\toprule
\textbf{Model} & \textbf{Setting}  & \textbf{Align.} & \textbf{Sty.} & \textbf{Cont.} & \textbf{Char.} & \textbf{Qual.} \\
\toprule
\multirow{2}*{\textbf{ARLDM}} & \textbf{No Constraint} & 2.29 & 2.77 & 1.88 & 1.74 & 2.47 \\
                              & \textbf{LLM Constraints}& 3.22 & 3.38 & 2.49 & 2.08 & 3.02 \\
\midrule
\multirow{2}*{\textbf{MM-Inter.}} & \textbf{No Constraint} & 2.43 & 4.20 & 3.68 & 2.81 & 3.23 \\
                              & \textbf{LLM Constraints} & \textbf{3.32} & \textbf{4.49} & \textbf{4.03} & \textbf{2.95} & \textbf{3.42} \\
\midrule
\rowcolor{mygrey} \textbf{Gold Ref.}  & - & 4.68 & 4.89 & 4.82 & 4.72 & 4.80 \\
\bottomrule
\end{tabular}
}
\caption{Human evaluation results on VWP narratives, \wrt{} text-image alignment (\textbf{Align.}), style consistency (\textbf{Sty.}), content consistency (\textbf{Cont.}), character consistency (\textbf{Char.}) and image quality (\textbf{Qual.}). Best results (excluding Gold Ref.) are \textbf{bolded}.}
\label{tab:human_eval}
\end{table}

\begin{figure}[t]
\centering
\includegraphics[width=1.0\columnwidth]{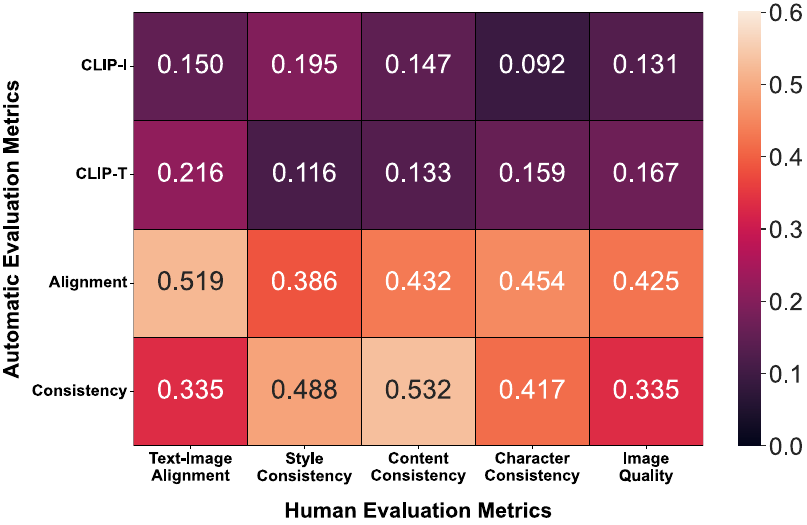}
\caption{Pearson correlation coefficients between human and automatic evaluation metrics on VWP narratives. \textbf{Alignment} and \textbf{Consistency} in automatic evaluation metrics denote the average of our VQA-based fine-grained alignment and consistency metrics, respectively, rooted on MiniCPM-V-2.6.}
\label{metric_corr}
\end{figure}

\begin{figure*}[t]
\centering
\includegraphics[width=0.9\textwidth]{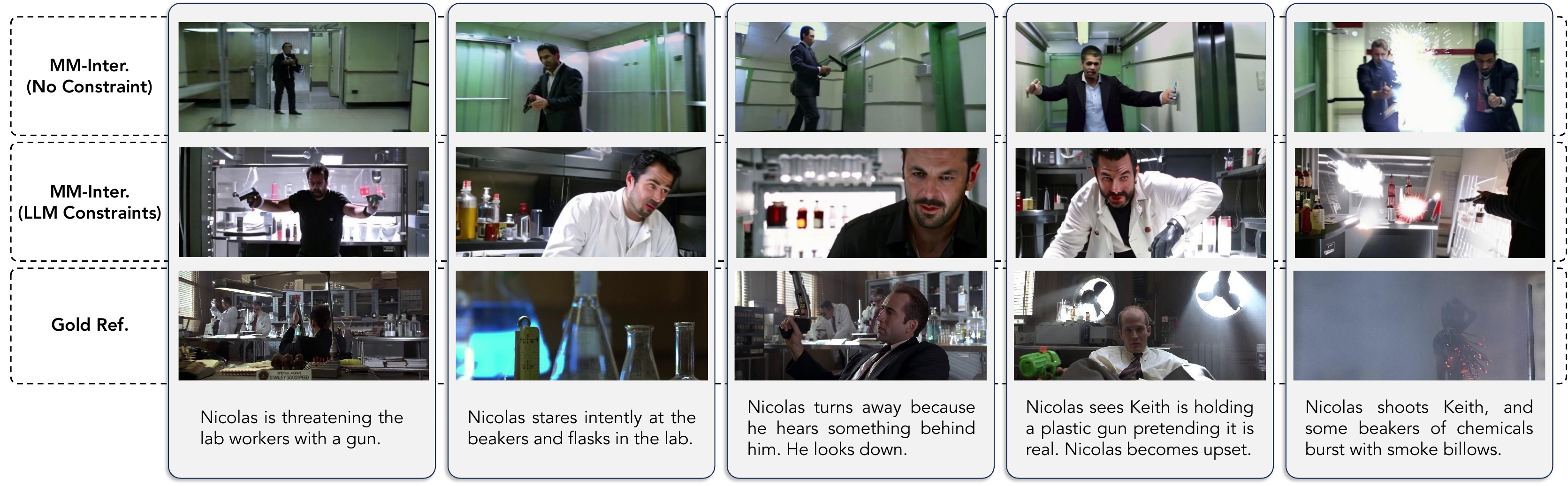}
\caption{Visual narratives generated by MM-Interleaved with and without LLM-generated narrative constraints, and the gold reference.}
\label{case_study}
\end{figure*}

Our human evaluation supports the results of our automatic evaluation.
12 expert annotators evaluate the visual narrative generations of ARLDM and MM-Interleaved models (with and without LLM generated constraints), along with the gold references, on 100 VWP testing samples.\footnote{Model generations and the gold reference are randomly shuffled in the human evaluation of each narrative sample, and human annotators are bling to the source of each (generated or reference) visual narrative.}
The annotators follow the methodology of StoryGen \citep{liu2024intelligent} and use a Likert scale from 1 to 5 (higher is better) to rate the visual narrative's \textbf{alignment} with input textual narrative, consistency of image \textbf{style}, non-character \textbf{content} and \textbf{character} appearance, and general image \textbf{quality}.
Our human evaluation results in Table~\ref{tab:human_eval} also validate that learning narrative constraints contributes to more faithful and consistent visual narratives.
Moreover, the human preference towards different vision models is coherent with the preference given by our proposed metrics, where ARLDM generations are in general comparable with MM-Interleaved in terms of the alignment with input textual narrative, but significantly fall behind MM-Interleaved in terms of consistency.
By contrast, FID, CLIP-I and CLIP-T scores show more preference to ARLDM than MM-Interleaved.

Figure~\ref{case_study} shows a case of visual narratives generated by MM-Interleaved, with and without narrative constraints. Compared to the model generation without constraints, the generation with LLM constraints includes more details that are faithful to the textual narrative, \ie{}, depicting a lab background and reasonable facial expressions of Nicolas according to the narrative.
Consistent with our human evaluation in Table~\ref{tab:human_eval}, however, the model generation significantly falls short of gold references \wrt{} character consistency, \eg{}, Nicolas' outfit shifts between black and white.

\paragraph{Reliability of Evaluation Metrics} We more closely study the correlation of our automatic evaluation metrics to the five human evaluation metrics.
In particular, we consider the average of our fine-grained alignment and consistency metrics, denoted as \textbf{Alignment} and \textbf{Consistency}, and compare them to the CLIP-based metrics CLIP-I and CLIP-T.
Using 100 VWP testing samples, we compute the Pearson correlation coefficient between human and automatic evaluation scores for four methods\footnote{We consider the four methods studied in the human evaluation, \ie{}, ARLDM with and without LLM constraints, and MM-Interleaved with and without LLM constraints.} and the gold references.
Figure~\ref{metric_corr} presents the results of our correlation study.
Compared to CLIP-I and CLIP-T, Alignment and Consistency metrics demonstrate overall better correlation with human evaluation, verifying that our proposed VQA-based evaluation gives more reliable results than CLIP-based similarity measure.


\begin{table}[t]
\centering
\resizebox{1.0\columnwidth}{!}{
\smallskip\begin{tabular}{@{~}l@{~~}l@{~~~~}c@{~~~~}c@{~~~~}c@{~}}
\toprule
\textbf{Model} & \textbf{Setting}  & \textbf{FID} & \textbf{Alignment} & \textbf{Consistency}\\
\toprule
\multirow{3}*{\textbf{ARLDM}} & \textbf{No Constraint} & 97.9 (55.4) & 0.295 (0.125) & 0.187 (0.220) \\
                              & \textbf{LLM Constraints} & \textbf{82.6} (45.0) & \textbf{0.479} (0.046) & 0.488 (0.210) \\
                              & \textbf{Gold Constraints} & \underline{77.7} (42.5) & \underline{0.566} (0.045) & 0.573 (0.135) \\
\midrule
\multirow{3}*{\textbf{StoryGen}} & \textbf{No Constraint} & 161.4 (82.8) & 0.227 (0.110) & 0.186 (0.074) \\
                              & \textbf{LLM Constraints} & 112.0 (59.9) & 0.375 (0.071) & 0.396 (0.049) \\
                              & \textbf{Gold Constraints} & 107.7 (58.7) & 0.457 (0.063) & 0.447 (0.028) \\
\midrule
\multirow{3}*{\textbf{MM-Inter.}} & \textbf{No Constraint} & 102.4 (54.1) & 0.276 (0.153) & 0.553 (0.106) \\
                              & \textbf{LLM Constraints} & 95.7 (53.5) & 0.466 (0.054) & \textbf{0.749} (0.060) \\
                              & \textbf{Gold Constraints} & 90.8 (51.5) & 0.556 (0.053) & \underline{0.797} (0.043) \\
\midrule
\rowcolor{mygrey} \textbf{Gold Ref.}  & - & - & 0.817 & 0.882 \\
\bottomrule
\end{tabular}
}
\caption{Zero-shot evaluation results on \textbf{Storyboard20K} narratives. All models are trained on VWP narratives. \textbf{Alignment} and \textbf{Consistency} denote the average score of our fine-grained alignment and consistency metrics rooted on MiniCPM-V-2.6. Performance drops compared to the results on VWP narratives are in brackets.
\colorbox{mygrey2}{Other notations are same as Table~\ref{tab:results_main}.}
}
\label{tab:generlize_sb20k}
\end{table}

\paragraph{Generalization}
For each model trained on VWP narratives, we also test its generalization performance to the Storyboard20K narratives in \ourbench{}.
We aggregate the generalization results in Table~\ref{tab:generlize_sb20k}, and compare them with the evaluation results on VWP testing samples.\footnote{Full results on Storyboard20K are in the supplementary material.}
Compared to \textit{No Constraint}, models augmented with narrative constraints yield smaller drops on all metrics when generalizing from VWP to Storyboard20K narratives, which indicates those models' more robust visual narrative capabilities on out-of-distribution samples, probably due to their learning of more generic visual narrative planning from the constraints.

\begin{table*}[t]
\centering
\resizebox{1.0\textwidth}{!}{
\smallskip\begin{tabular}{llcccccccccccc}
\toprule
\multirow{2}*{\textbf{Model}} & \multirow{2}*{\textbf{Setting}} & \multirow{2}*{\textbf{FID}}  & \multirow{2}*{\textbf{CLIP-I}} & \multirow{2}*{\textbf{CLIP-T}} & \multirow{2}*{\textbf{CLIP-T}} & \multicolumn{5}{c}{\textbf{Alignment}} & \multicolumn{3}{c}{\textbf{Consistency}} \\
    \cmidrule(lr){7-11}  \cmidrule(lr){12-14} 
 &  &  &  &  & \textbf{-MRR} & \textbf{Ent.} & \textbf{Num.} & \textbf{Attr.} & \textbf{Time} & \textbf{Loc.} & \textbf{Sty.} & \textbf{Char.} & \textbf{Loc.} \\
\toprule
\multirow{3}*{\textbf{ARLDM}} & \textbf{No Constraint} & 64.7 & 0.628 & 0.198 & 0.102 & 0.471 & 0.288 & 0.167 & 0.405 & 0.380 & 0.500 & 0.146 & 0.290 \\
                              & \textbf{LLM Constraints} & 56.7 & 0.652 & 0.200 & 0.143 & \textbf{0.569} & 0.307 & 0.222 & 0.441 & 0.442 & 0.656 & 0.262 & 0.330\\
                              & \textbf{Gold Constraints} & 56.5 & 0.689 & \underline{0.202} & \underline{0.149} & \underline{0.577} & 0.357 & 0.268 & 0.489 & 0.486 & 0.688 & 0.289 & 0.384 \\
\midrule
\multirow{3}*{\textbf{StoryGen}} & \textbf{No Constraint} & 63.6 & 0.646 & 0.195 & 0.101 & 0.454 & 0.283 & 0.165 & 0.397 & 0.374 & 0.425 & 0.104 & 0.227 \\
                              & \textbf{LLM Constraints} & \textbf{56.2} & \textbf{0.660} & \textbf{0.201} & \textbf{0.144} & 0.563 & 0.307 & 0.208 & 0.423 & 0.397 & 0.607 & 0.289 & 0.319 \\
                              & \textbf{Gold Constraints} & \underline{55.6} & \underline{0.692} & 0.202 & 0.147 & 0.571 & 0.352 & 0.258 & 0.469 & 0.439 & 0.647 & 0.291 & 0.375 \\
\midrule
\multirow{3}*{\textbf{MM-Inter.}} & \textbf{No Constraint} & 74.9 & 0.637 & 0.183 & 0.058 & 0.448 & 0.292 & 0.184 & 0.423 & 0.374 & 0.945 & 0.335 & 0.702 \\
                              & \textbf{LLM Constraints} & 72.9 & 0.655 & 0.188 & 0.107 & 0.535 & \textbf{0.366} & \textbf{0.265} & \textbf{0.473} & \textbf{0.459} & \textbf{0.956} & \textbf{0.355} & \textbf{0.780} \\
                              & \textbf{Gold Constraints} & 72.0 & 0.678 & 0.190 & 0.112 & 0.545 & \underline{0.413} & \underline{0.313} & \underline{0.503} & \underline{0.561} & \underline{0.969} & \underline{0.383} & \underline{0.803} \\
\midrule
\rowcolor{mygrey} \textbf{Gold Ref.}  & - & - & - & 0.207 & 0.160 & 0.758 & 0.817 & 0.778 & 0.755 & 0.752 & 0.969 & 0.769 & 0.814 \\
\bottomrule
\end{tabular}
}
\caption{Evaluation results on \textbf{StorySalon} narratives. \colorbox{mygrey2}{Notations are same as Table~\ref{tab:results_main}.}}
\label{tab:results_main_ssalon}
\end{table*}

\paragraph{Ablation Study}
In Table~\ref{tab:results_main}, we also conduct ablation study on the using MM-Interleaved with LLM Constraints, to more finely investigate the benefits of adding commonsense links and discourse features as visual narrative constraints.
Specifically, we individually remove the commonsense links inserted in the image caption (\textbf{w/o CL}), the whole serialized discourse features (\textbf{w/o DF}), the global discourse features (\textbf{w/o GDF}) or the scene-specific discourse features (\textbf{w/o SDF}), from the LLM-generated constraints, and re-train the vision model to generate the visual narrative based on the textual narrative and ablated constraints.
Our results show that removing either commonsense links or any subset of the discourse features leads to performance degradation on all metrics, indicating that both commonsense and discourse constraints provide complementary benefits for visual narrative generation.
However, we detect greater degradation of w/o DF compared to w/o CL on most metrics, revealing that discourse constraints may be more beneficial for improving visual narratives, especially \wrt{} generating the location and non-character contents where significant gaps between w/o CL and w/o DF are found.

One concern of augmenting the vision model with narrative constraints is whether the improvements are just due to adding more input text tokens.
To eliminate this concern, we include another ablation study (\textbf{Random}), where we group the training narrative samples by their length (\ie{}, number of scenes or images), randomly shuffle the constraints of narrative samples in the same group, and use the shuffled samples to re-train the vision model.
Results of \textbf{Random} are worse than the \textbf{No Constraint} setting, showing that generated visual narratives only benefit from aligned narrative constraints, and not random ones.

\vspace{-5pt}
\paragraph{Correlation between Visual Generation and Constraints}
We further analyze how the alignment of narrative constraints and textual narrative affects the faithfulness of visual narrative generation.
For each scene in the VWP testing samples, we calculate the CLIP text embedding similarity between the scene's textual narrative and serialized narrative constraints, and pair it with the CLIP text-image embedding similarity between the scene's textual narrative and visual narrative image generated by MM-Interleaved.\footnote{The serialized narrative constraints are either from gold labels (in which case the corresponding CLIP text-image similarity is computed using images generated with Gold Constraints) or LLM-generated (in which case the corresponding CLIP text-image similarity is computed using images generated with LLM Constraints)}
Our paired similarity scores achieve $\sim$0.4 Pearson correlation coefficient on both Gold and LLM settings,\footnote{We include the visualization of the paired similarity score distribution in the supplementary material.} 
indicating a clear positive correlation between (a) the alignment of a textual narrative  and its constraints, and (b) the alignment between the same textual narrative and its visual manifestation. This finding highlights the significance of planning intermediate constraints to promote faithful visual narratives.

\subsection{StorySalon Narratives}
Table~\ref{tab:results_main_ssalon} presents the evaluation results of our deployed baseline methods on \ourbench{}'s StorySalon narratives.
We draw the same conclusion as on VWP and Storyboard20K narratives that incorporating narrative constraints effectively improves the faithfulness and self-consistency of visual narrative generation.
The coherent results on all types of \ourbench{} narratives imply the ubiquity of implicit commonsense and discourse constraints in visual narratives, which also indicate that our proposed knowledge augmentation framework is universally effective on various visual narrative domains and image styles.

\section{Conclusion}
\label{sec:conclusion}

In this work, we propose a new benchmark \ourbench{} that draws attention to the faithfulness and self-consistency challenges of visual narrative generation.
\ourbench{} provides a reliable foundation for generative vision models to learn faithful visual narratives with discourse and commonsense constraints.
In view of the shortcomings of visual narrative evaluation, \ourbench{} also proposes new metrics to more closely assess the consistency of visual narrative generations and their alignment with the input textual narrative.
Our results indicate that model-generated visual narratives have considerable room for improvement to reach the level of human visual storytelling, which calls for future study on more robust visual narrative generators.

\section*{Acknowledgements}
We gratefully acknowledge the support of the Swiss National Science Foundation (No. 215390), Innosuisse (PFFS-21-29), the EPFL Center for Imaging, Sony Group Corporation, and a Meta LLM Evaluation Research Grant.

{
    \small
    \bibliographystyle{ieeenat_fullname}
    \bibliography{main}
}

\clearpage

\clearpage
\setcounter{page}{1}
\maketitlesupplementary

\setcounter{section}{0}
\renewcommand{\thesection}{S\arabic{section}}
\setcounter{table}{0}
\renewcommand{\thetable}{S\arabic{table}}
\setcounter{figure}{0}
\renewcommand{\thefigure}{S\arabic{figure}}

The supplementary materials contain the following information and materials:
\begin{itemize}
    \item Data construction details (Section~\ref{sec:data_details}).
    \item Evaluation details (Section~\ref{sec:eval_details}).
    \item Experimental setup details (Section~\ref{sec:exp_details}).
    \item Full experimental results (Section~\ref{sec:full_results})
\end{itemize}


\section{\ourbench{} Data Construction Details}
\label{sec:data_details}
The visual-textual narrative pairs in our benchmark are sampled from three diverse visual storytelling datasets, including Visual Writing Prompts (VWP) \citep{hong2023visual}, Storyboard20K \citep{xie2024learning} and StorySalon \cite{liu2024intelligent}.
The VWP dataset contains $\sim$12K narrative samples, whose visual narrative scenes are extracted and curated from MovieNet \citep{huang2020movienet} frames, with corresponding textual narratives crafted by Amazon Mechanical Turk (AMT) workers.
The Storyboard20K dataset covers a broader set of visual narrative scenes sampled from MovieNet and also LSMDC \citep{rohrbach2017movie}, with real movie synopses collected by a two-stage approach of automatic tagging and manual calibration.
We filter the narrative samples in Storyboard20K to keep $\sim$10K of them, which have aligned shot-by-shot movie synopses, serving as the textual narratives.
Different from the movie-based narratives in VWP and Storyboard20K, the StorySalon dataset is oriented to animation-style visual narratives, whose images and aligned narrative texts are extracted from diverse YouTube videos and E-books.
We use the Google Translation API\footnote{\url{https://github.com/ssut/py-googletrans}} to translate non-English narrative texts collected in StorySalon into English.
To ensure accurate translation, we only apply the API to $\sim$26K StorySalon scenes (or images) whose associated narrative texts are in the 19 common languages shown in Table~\ref{tab:salon_trans_lang}, and then exclude the narrative samples whose texts are not fully translated into English.
Besides, we filter the StorySalon samples whose textual narratives are poor-annotated, \ie{}, $>$10\% of the sample's scenes are annotated with uninformative texts containing less than 5 words.
Finally, $\sim$2K narrative samples from StorySalon are included.

\begin{figure*}[t]
\centering
\includegraphics[width=0.6\textwidth]{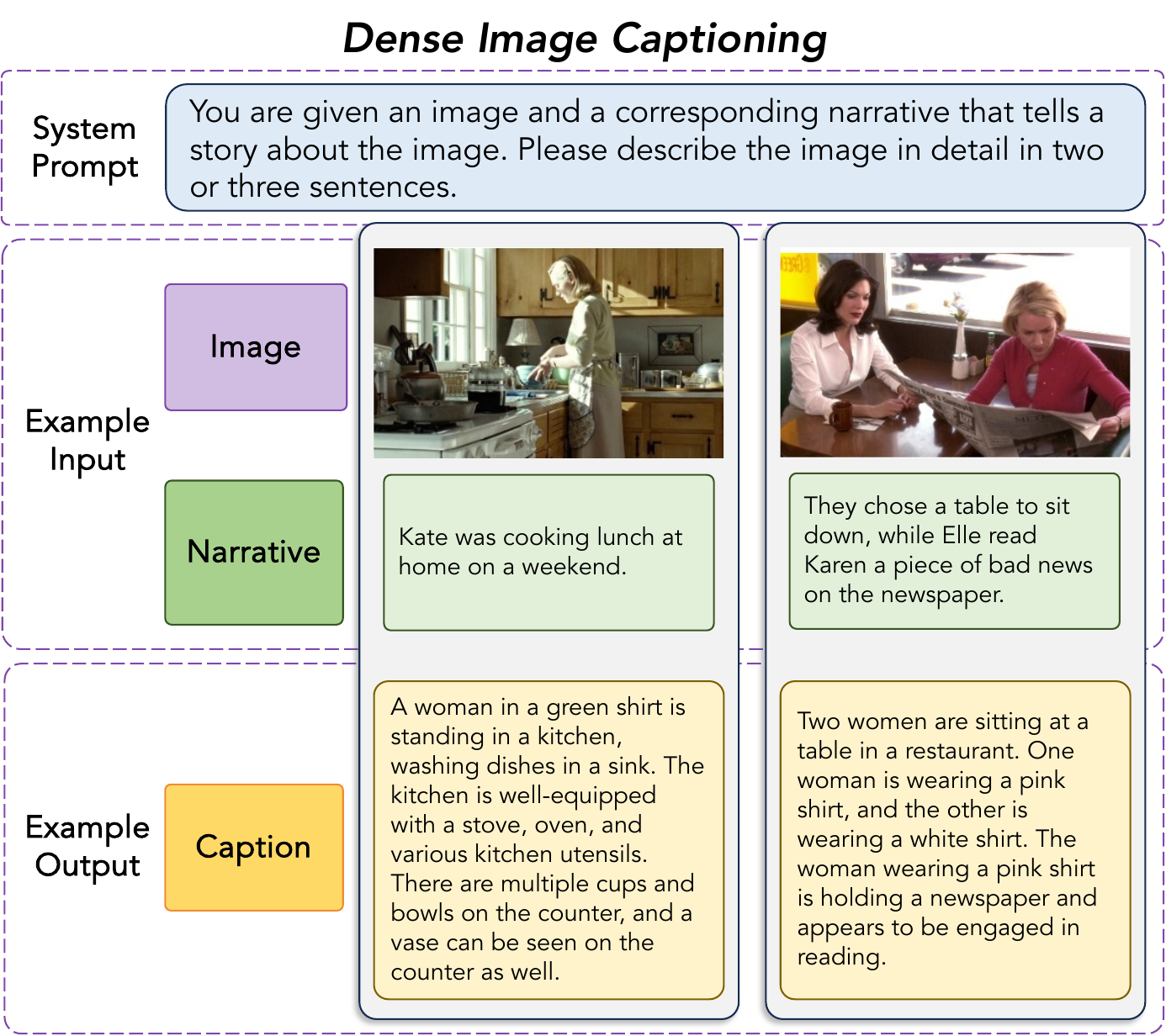}
\caption{Few-shot prompting demonstrations for constructing the dense \textbf{image captions} in \ourbench{}.}
\label{dense_caption}
\end{figure*}

\begin{figure*}[t]
\begin{subfigure}{1.0\textwidth}
  \centering
  \includegraphics[width=0.92\linewidth]{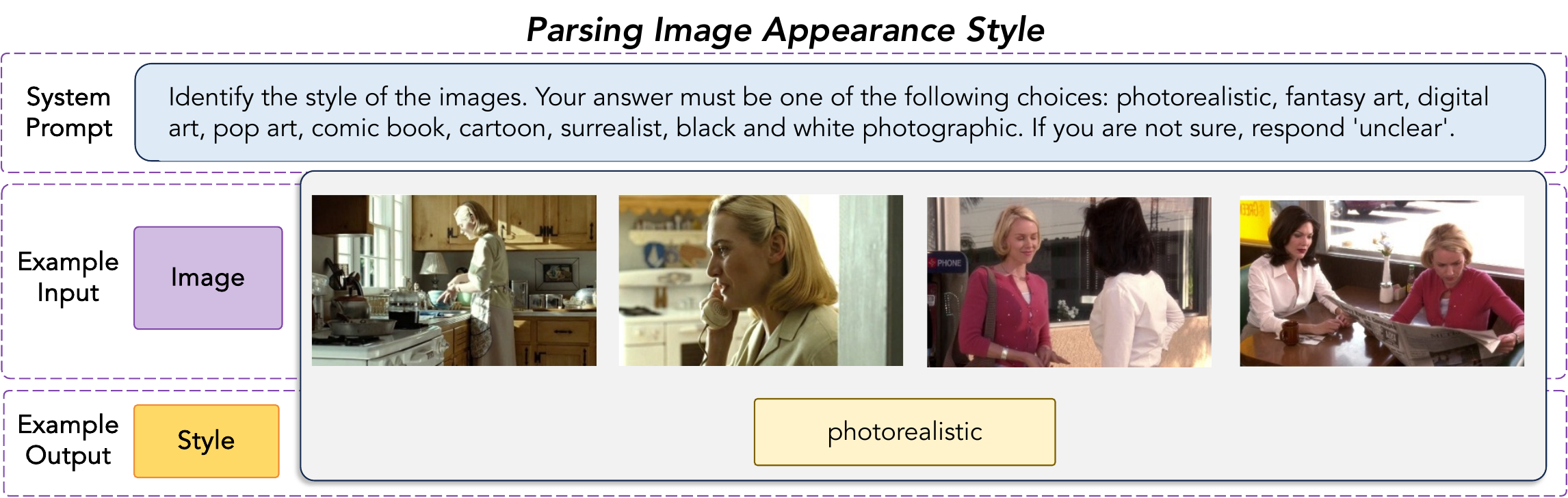}
  \caption{}
  \label{global_features_1}
\end{subfigure}
\begin{subfigure}{1.0\textwidth}
  \centering
  \includegraphics[width=0.9\linewidth]{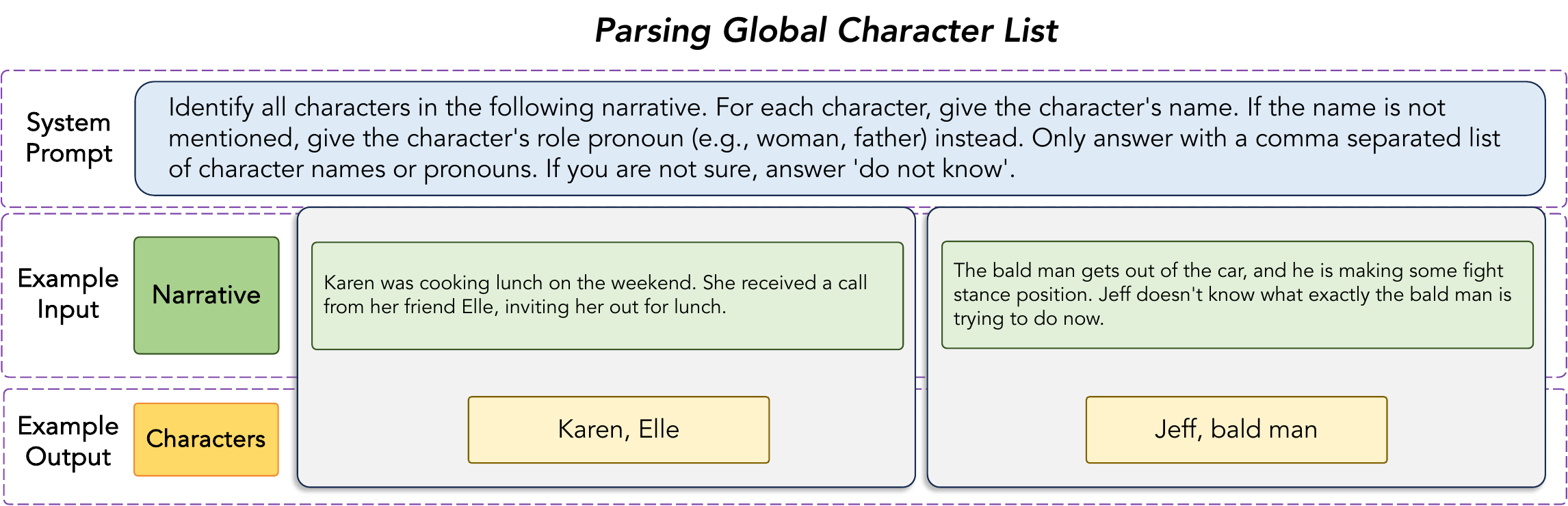}
  \caption{}
  \label{global_features_2}
\end{subfigure}
\begin{subfigure}{1.0\textwidth}
  \centering
  \includegraphics[width=0.76\linewidth]{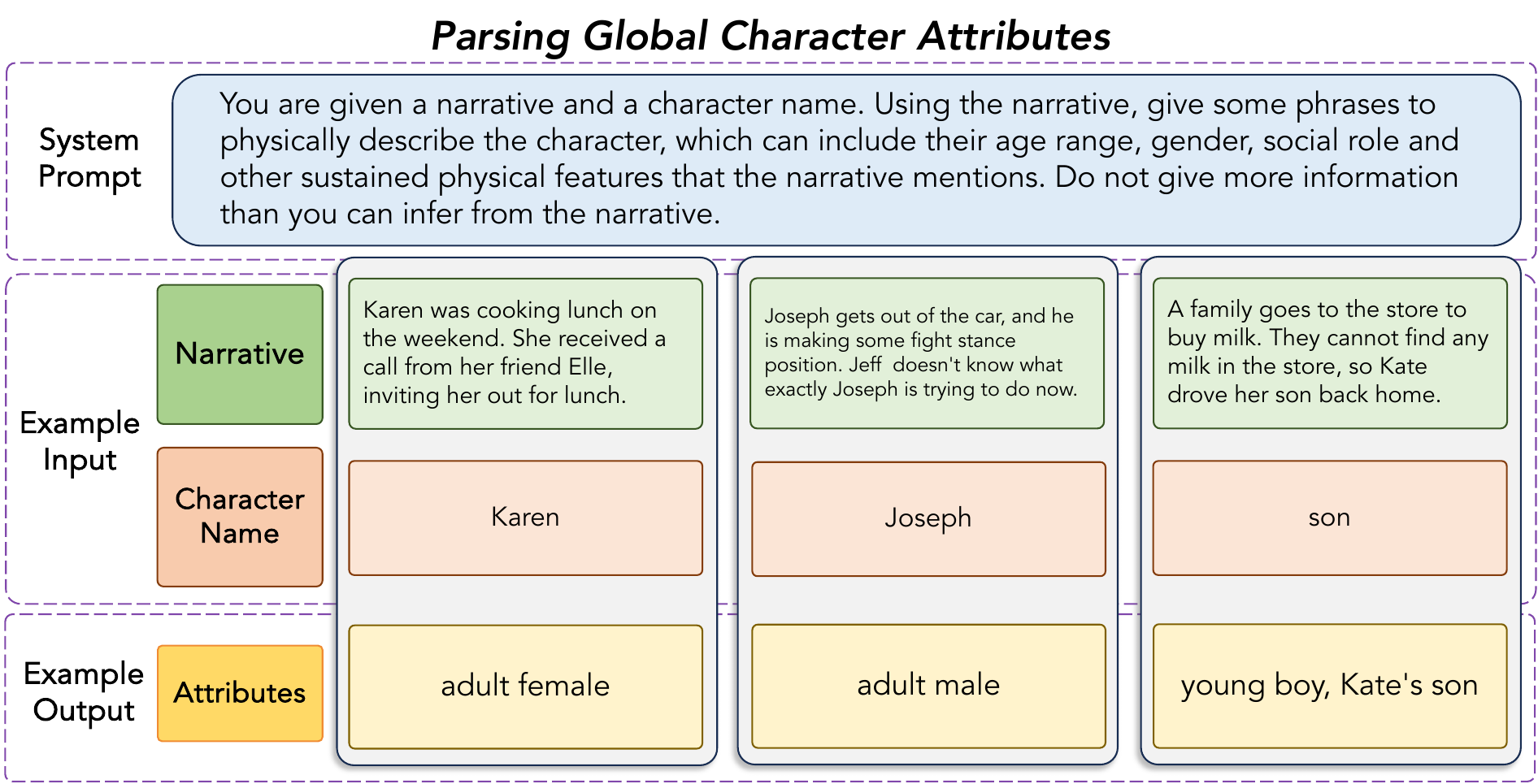}
  \caption{}
  \label{global_features_3}
\end{subfigure}
\caption{Few-shot prompting demonstrations for parsing the \textbf{global features} in \ourbench{}, including (a) image appearance style, (b) character list, and (c) character attributes. The output features of (b) and (c) form the global profile of characters.}
\label{global_features}
\end{figure*}

Based on the sampled visual-textual narrative pairs, \ourbench{} further annotates the commonsense and discourse constraints underlying each narrative sample, by prompting advanced VLMs and LLMs instead of relying on human annotators.
Table~\ref{tab:few_shot_num} summarizes the number of few-shot prompting examples used for each step of our \ourbench{} constraint annotation.
For each annotation step, we tune the number of few-shot examples on a scale of $1$ to $3$, and select the number that leads to the best annotation results in our pilot study on 10 narrative samples.
Figure~\ref{dense_caption}~-~\ref{scene_features} list the specific few-shot examples and instructions that we finally used for annotating the image captions, commonsense links, global and scene features in \ourbench{}, respectively.

\begin{table}[t]
\centering
\resizebox{0.85\columnwidth}{!}{
\smallskip\begin{tabular}{lclc}
\toprule
\textbf{Language} & \textbf{\# Scenes} & \textbf{Language} & \textbf{\# Scenes} \\
\toprule
Hindi (hi) & 8213 & Hausa (ha) & 926 \\
French (fr) & 2503 & Spanish (es) & 758 \\
Indonesian (id) & 2197 & Italian (it) & 386 \\
Arabic (ar) & 2053 & Dutch (nl) & 198 \\
Marathi (mr) & 1544 & German (de) & 187 \\
Nepali (ne) & 1521 & Portuguese (pt) & 137 \\
Afrikaans (af) & 1464 & Finnish (fi) & 113 \\
Swahili (sw) & 1311 & Welsh (cy) & 82 \\
Vietnamese (vi) & 1220 & Polish (pl) & 78 \\
Uzbek (uz) & 1150 & \textbf{Total} & \textbf{26041} \\
\bottomrule
\end{tabular}
}
\caption{Statistics of StorySalon scenes (or images) whose associated non-English narrative texts are translated into English.}
\label{tab:salon_trans_lang}
\end{table}

\begin{table}[t]
\centering
\resizebox{1.0\columnwidth}{!}{
\smallskip\begin{tabular}{cccccccccc}
\toprule
\textbf{Cap.} & \textbf{Ent.} & \textbf{CL} & \textbf{Sty.} & \textbf{List} & \textbf{Attr.} & \textbf{Num.} & \textbf{Name} & \textbf{Time} & \textbf{Loc.}\\
\toprule
 2 & 3 & 3 & 1 & 2 & 3 & 3 & 2 & 3 & 2 \\
\bottomrule
\end{tabular}
}
\caption{Number of few-shot examples used for \ourbench{} data annotation, including dense image captioning (\textbf{Cap.}), visual entity extraction from dense captions (\textbf{Ent.}), commonsense link construction (\textbf{CL}), and the parsing of image appearance style (\textbf{Sty.}), global character list (\textbf{List}) and attributes (\textbf{Attr.}), and each scene's presented character number (\textbf{Num.}) and name (\textbf{Name}), time of day (\textbf{Time}) and location (\textbf{Loc.}).}
\label{tab:few_shot_num}
\end{table}

\begin{table}[t]
\centering
\resizebox{1.0\columnwidth}{!}{
\smallskip\begin{tabular}{llcccccccc}
\toprule
\multirow{2}*{\textbf{Source}} & \multirow{2}*{\textbf{Set}} & \multirow{2}*{\textbf{\# Nar.}} & \multirow{2}*{\textbf{\# Sce.}} & \multicolumn{2}{c}{\textbf{Avg. \# Char.}} & \multirow{2}*{\textbf{\# CL}} & \multicolumn{3}{c}{\textbf{\# Label Types}}\\
    \cmidrule(lr){5-6}  \cmidrule(lr){8-10}
 &  &  &  & \textbf{per Nar.} & \textbf{per Sce.} &  & \textbf{Sty.} & \textbf{Time} & \textbf{Loc.} \\
\toprule
\multirow{2}*{\textbf{VWP}} & train & 11652 & 66632 & 3.07 & 1.71 & 274861 & \multirow{2}*{9} & \multirow{2}*{6} & 545 \\
                            & test & 834 & 4901 & 3.00 & 1.73 & 22434 &  &  & 191 \\
\midrule
\multirow{1}*{\textbf{Story-}} & train & 9252 & 92520 & 3.73 & 1.40 & 316356 & \multirow{2}*{9} & \multirow{2}*{6} & 551 \\
\multirow{1}*{\textbf{board20K}} & test & 1194 & 11940 & 4.07 & 1.59 & 40576 &  &  & 341 \\
\midrule
\multirow{1}*{\textbf{Story-}} & train & 1593 & 21827 & 6.49 & 2.14 & 71489 & \multirow{2}*{9} & \multirow{2}*{6} & 518 \\
\multirow{1}*{\textbf{Salon}} & test & 85 & 1181 & 6.64 & 2.03 & 4374 &  &  & 111 \\
\bottomrule
\end{tabular}
}
\caption{Statistics of \ourbench{} data samples and annotations, including total number of narratives (\textbf{\# Nar.}), total number of scenes or images (\textbf{\# Sce.}), average number of distinct characters per narrative (\textbf{Avg. \# Char. per Nar.}), average number of presented characters per scene (\textbf{Avg. \# Char. per Sce.}), total number of commonsense links (\textbf{\# CL}), total types of appearance style (\textbf{\# Sty.}), time of day (\textbf{\# Time}) and location (\textbf{\# Loc.}) labels.}
\label{tab:data_stats}
\end{table}

\begin{table*}[t]
\centering
\resizebox{1.0\textwidth}{!}{
\smallskip\begin{tabular}{lll}
\toprule
\textbf{Aspect} & \textbf{Metric} & \textbf{Demonstration}\\
\toprule
\multirow{15}*{\textbf{Alignment}} & \multirow{2}*{\textbf{Non-Character}} & \{\textit{generated image for a scene}\} \\
                                  &  & Does this image contain or imply \{\textit{each non-character visual entity in the scene's gold commonsense links}\}? Only answer yes or no. \\
                                    \cmidrule(lr){2-3}
                                  & \multirow{2}*{\textbf{Character Number}} & \{\textit{generated image for a scene}\} \\
                                  &  & How many characters are in this image? Only answer an Arabic number. \\
                                    \cmidrule(lr){2-3}
                                  & \multirow{6}*{\textbf{Character Attribute}} & \{\textit{generated image for a scene}\} \\
                                  &  & Character descriptions: \\
                                  &  & \{\textit{gold character 1 presented in the scene features}\}: \{\textit{profile of character 1 in the global features}\}\\
                                  &  & \{\textit{gold character 2 presented in the scene features}\}: \{\textit{profile of character 2 in the global features}\}\\
                                  &  & ... \\
                                  &  & Do characters in this image fit into their descriptions? Only answer yes or no. \\
                                    \cmidrule(lr){2-3}
                                  & \multirow{2}*{\textbf{Time of Day}} & \{\textit{generated image for a scene}\} \\
                                  &  & Is this image taken in (or at) the \{\textit{gold time of day labeled in the scene features}\}? Only answer yes or no. \\
                                    \cmidrule(lr){2-3}
                                  & \multirow{2}*{\textbf{Location}} & \{\textit{generated image for a scene}\} \\
                                  &  & Is this image taken at a (or an) \{\textit{gold location labeled in the scene features}\}? Only answer yes or no. \\
\midrule
\multirow{9}*{\textbf{Consistency}} & \multirow{2}*{\textbf{Style}} & \{\textit{generated image for scene 1}\} \{\textit{generated image for scene 2}\} ...  \{\textit{generated image for scene N}\}\\
                                    &  & Are all these images in the same style? Only answer yes or no. \\
                                    \cmidrule(lr){2-3}
                                    & \multirow{3}*{\textbf{Character}} & \{\textit{generated image for scene X}\} \{\textit{generated image for scene Y}\} ...\\
                                    &  & Do all these images contain the same character \{\textit{each overlapped character across the scenes X, Y, ..., indicated by their scene features}\}:\\
                                    &  & \{\textit{profile of the overlapped character in the global features}\}? Only answer yes or no. \\
                                    \cmidrule(lr){2-3}
                                    & \multirow{3}*{\textbf{Location}} & \{\textit{generated image for scene X}\} \{\textit{generated image for scene Y}\} ...\\
                                    &  & Are all these images taken at the same \{\textit{gold location label shared by the scenes X, Y, ..., indicated by their scene features}\}? \\
                                    &  & Only answer yes or no.\\
\bottomrule
\end{tabular}
}
\caption{VQA demonstrations used for the fine-grained alignment and consistency metrics in \ourbench{}. For Alignment of Character Number, we record the average probability of the VLMs (MiniCPM-V-2.6 or LLaVA-OneVision-72B) outputting the correct character number as its first decoded token (or if characters are more than 9, the same number of leading tokens as the correct number of digits). For other metrics, we report the average probability of the VLM outputting \textit{Yes} as its first decoded token. The spans labeled by ``\{\}'' in the demonstrations are replaced by their corresponding texts or images.}
\label{tab:eval_prompt}
\end{table*}

According to \ourbench{} annotations, we also exclude the narrative samples that contain no character or commonsense link.
Table~\ref{tab:data_stats} shows the final statistics of \ourbench{} narrative samples and annotations.
Each \ourbench{} narrative sample contains $\sim$8.09 scenes (or images) in average, which is longer than prior image sequences (with a length of $5$) studied in visual narrative generation, \ie{}, VIST \citep{huang2016visual}, PororoSV \citep{li2019storygan} and FlintstonesSV \cite{maharana2021integrating}.
Besides, \ourbench{} incorporates new annotations of fine-grained visual narrative constraints, which are not involved in previous visual narrative studies.

\section{\ourbench{} Evaluation Details}
\label{sec:eval_details}

We adopt zero-shot prompting to implement all of our proposed VQA-based fine-grained alignment and consistency metrics in \ourbench{}.
Table~\ref{tab:eval_prompt} lists the specific demonstrations used for our VQA-based metrics.
The VQA score of non-character alignment metric is averaged across each non-character visual entity labeled in gold commonsense links.
While for other fine-grained alignment metrics, we calculate the average VQA score across each scene in the testing narrative samples.
For the style consistency metric, since it is based on all scenes of a narrative, we average the VQA score across each testing narrative sample.
In terms of the character and location consistency metrics, the VQA score is averaged across each gold character or location labeled in the narrative that is shared by multiple scenes.

\section{Experimental Setup Details}
\label{sec:exp_details}

For the setting of training visual narrative models with LLM Constraints, we preprocess our annotated commonsense and discourse constraints in \ourbench{}, to enable training the auto-regressive LLM (Llama3.1-70B-Instruct \citep{dubey2024llama}) to generate those constraints.
First, we merge the commonsense links into the dense image caption.
Specifically, for each entity in the image caption, if it appears in one of the commonsense links, we insert its linked textual narrative phrase right after the entity (in parentheses).
For example, if the image caption is \textit{A woman wearing a green shirt}, and its entity \textit{woman} is linked to the character \textit{Samantha} in the textual narrative, the caption will be converted to \textit{A woman (Samantha) wearing a green shirt}.
Second, we use a template to serialize the scene features, and insert presented characters' attributes in the global features.
For instance, if the scene features indicate that the presented character, time of day and location are \textit{Samantha}, \textit{afternoon} and \textit{kitchen}, respectively, and \textit{Samantha} has the profile \textit{adult female, wife} in the global features, the scene features will be serialized into the text sequence: \textit{[Characters] Samantha (adult female, wife) [Time of Day] afternoon [Location] kitchen}.
We train the LLM to auto-regressively generate the concatenation of image caption (with commonsense links inserted) and serialized scene features, as the narrative constraints used for augmenting the visual narrative generation.

\begin{table*}[t]
\centering
\resizebox{1.0\textwidth}{!}{
\smallskip\begin{tabular}{l@{~~}lc@{~~}cc@{~~}cc@{~~}cc@{~~}cc@{~~}cc@{~~}c}
\toprule
\multirow{2}*{\textbf{Model}} & \multirow{2}*{\textbf{Setting}}  & \multicolumn{2}{c}{\textbf{Ranking}} & \multicolumn{2}{c}{\textbf{Non-Character}} & \multicolumn{2}{c}{\textbf{Character Number}} & \multicolumn{2}{c}{\textbf{Character Attribute}} & \multicolumn{2}{c}{\textbf{Time of Day}} & \multicolumn{2}{c}{\textbf{Location}}\\
    \cmidrule(lr){3-4}  \cmidrule(lr){5-6}  \cmidrule(lr){7-8}  \cmidrule(lr){9-10}  \cmidrule(lr){11-12}  \cmidrule(lr){13-14}
 &  & \textbf{CLIP-T-MRR} & \textbf{VQA-MRR} & \textbf{MiniCPM} & \textbf{Llava} & \textbf{MiniCPM} & \textbf{Llava} & \textbf{MiniCPM} & \textbf{Llava} & \textbf{MiniCPM} & \textbf{Llava} & \textbf{MiniCPM} & \textbf{Llava} \\
\toprule
\multirow{3}*{\textbf{ARLDM}} & No Constraint & 0.1096 & 0.1435 & 0.5640 & 0.5419 & 0.3980 & 0.3858 & 0.3199 & 0.3176 & 0.4429 & 0.3942 & 0.3759 & 0.4031 \\
                              & LLM Constraints & \textbf{0.1508} & \textbf{0.2423} & \textbf{0.6741} & \textbf{0.6344} & 0.4434 & 0.4345 & 0.4107 & 0.3785 & \textbf{0.5119} & \textbf{0.4810} & 0.5835 & 0.5825 \\
                              & Gold Constraints & \underline{0.1551} & \underline{0.2503} & \underline{0.6823} & \underline{0.6420} & 0.6188 & 0.5607 & \underline{0.5464} & 0.5573 & \underline{0.5183} & \underline{0.4945} & 0.6899 & 0.5650 \\
\midrule
\multirow{3}*{\textbf{StoryGen}} & No Constraint & 0.1003 & 0.1158 & 0.4708 & 0.4707 & 0.3352 & 0.3236 & 0.2846 & 0.2167 & 0.2788 & 0.2804 & 0.3153 & 0.3791 \\
                              & LLM Constraints & 0.1056 & 0.1503 & 0.5950 & 0.5764 & 0.4236 & 0.4028 & 0.3412 & 0.3191 & 0.3673 & 0.3444 & 0.5041 & 0.5460 \\
                              & Gold Constraints & 0.1151 & 0.1728 & 0.6138 & 0.5873 & 0.5474 & 0.5081 & 0.4443 & 0.3749 & 0.3930 & 0.3467 & 0.5982 & 0.6325 \\
\midrule
\multirow{6}*{\textbf{MM-Inter.}} & No Constraint & 0.0660 & 0.1126 & 0.4990 & 0.4856 & 0.4088 & 0.3982 & 0.3259 & 0.3265 & 0.4632 & 0.4373 & 0.4489 & 0.4713 \\
                              & LLM Constraints & 0.1107 & 0.2074 & 0.6434 & 0.5942 & \textbf{0.4578} & \textbf{0.4407} & \textbf{0.4118} & \textbf{0.3915} & 0.4856 & 0.4745 & \textbf{0.5998} & \textbf{0.6016} \\
                              & \quad - w/o CL & 0.1090 & 0.2037 & 0.6422 & 0.5934 & 0.4546 & 0.4344 & 0.4092 & 0.3870 & 0.4748 & 0.4681 & 0.5968 & 0.5944 \\
                              & \quad - w/o DS & 0.1074 & 0.1983 & 0.6238 & 0.5872 & 0.4489 & 0.4355 & 0.4005 & 0.3887 & 0.4742 & 0.4635 & 0.5642 & 0.5734 \\
                              & \quad - Random & 0.0476 & 0.0861 & 0.4149 & 0.4152 & 0.3986 & 0.3904 & 0.3180 & 0.3135 & 0.4120 & 0.3849 & 0.4116 & 0.4335 \\
                              & Gold Constraints & 0.1179 & 0.2105 & 0.6521 & 0.6054 & \underline{0.6226} & \underline{0.5634} & 0.5462 & \underline{0.5736} & 0.4965 & 0.4841 & \underline{0.7276} & \underline{0.7157} \\
\midrule
\rowcolor{mygrey} \textbf{Gold Ref.}  & - & 0.1586 & 0.2662 & 0.7755 & 0.7163 & 0.8127 & 0.7652 & 0.7581 & 0.7157 & 0.7555 & 0.7196 & 0.8632 & 0.8100 \\
\bottomrule
\end{tabular}
}
\caption{Full evaluation results of our ranking-based and fine-grained \textbf{Alignment} metrics on VWP narratives. \textit{MiniCPM} and \textit{Llava} denote our fine-grained VQA-based metrics deployed on MiniCPM-V-2.6 and LLaVA-OneVision-72B. \textit{Gold Ref.} denotes gold references. \colorbox{mygrey2}{Best results with \textit{LLM Constraints} and with \textit{Gold Constraints} are \textbf{bolded} and \underline{underlined}, respectively.}}
\label{tab:results_alignment_vwp_full}
\end{table*}

\begin{table}[t]
\centering
\resizebox{1.0\columnwidth}{!}{
\smallskip\begin{tabular}{@{~}l@{~~}lc@{~~}cc@{~~}cc@{~~}c@{~}}
\toprule
\multirow{2}*{\textbf{Model}} & \multirow{2}*{\textbf{Setting}}  & \multicolumn{2}{c}{\textbf{Style}} & \multicolumn{2}{c}{\textbf{Character}} & \multicolumn{2}{c}{\textbf{Location}}\\
    \cmidrule(lr){3-4}  \cmidrule(lr){5-6}  \cmidrule(lr){7-8}
 &  & \textbf{MiniCPM} & \textbf{Llava} & \textbf{MiniCPM} & \textbf{Llava} & \textbf{MiniCPM} & \textbf{Llava} \\
\toprule
\multirow{3}*{\textbf{ARLDM}} & No Constraint & 0.4664 & 0.5857 & 0.3793 & 0.4102 & 0.3759 & 0.1788 \\
                              & LLM Constraints & 0.8586 & 0.7434 & 0.5507 & 0.5215 & 0.6888 & 0.3472 \\
                              & Gold Constraints & 0.8539 & 0.7326 & 0.5687 & 0.5280 & 0.6972 & 0.4359 \\
\midrule
\multirow{3}*{\textbf{StoryGen}} & No Constraint & 0.2379 & 0.4936 & 0.2305 & 0.3809 & 0.3106 & 0.2020 \\
                              & LLM Constraints & 0.4523 & 0.5390 & 0.4177 & 0.5014 & 0.4649 & 0.3192 \\
                              & Gold Constraints & 0.4747 & 0.5421 & 0.4233 & 0.5105 & 0.5272 & 0.3800 \\
\midrule
\multirow{6}*{\textbf{MM-Inter.}} & No Constraint & 0.9470 & 0.8077 & 0.5823 & 0.5631 & 0.4489 & 0.4831 \\
                              & LLM Constraints & \textbf{0.9859} & \textbf{0.8672} & \textbf{0.6780} & \textbf{0.6375} & \textbf{0.7642} & \textbf{0.6151} \\
                              & \quad - w/o CL & 0.9829 & 0.8664 & 0.6431 & 0.6290 & 0.7577 & 0.6113 \\
                              & \quad - w/o DS & 0.9776 & 0.8604 & 0.6443 & 0.6095 & 0.6842 & 0.5937 \\
                              & \quad - Random & 0.9453 & 0.7933 & 0.5763 & 0.5768 & 0.4471 & 0.4769 \\
                              & Gold Constraints & \underline{0.9764} & \underline{0.8542} & \underline{0.6880} & \underline{0.6399} & \underline{0.8558} & \underline{0.6931} \\
\midrule
\rowcolor{mygrey} \textbf{Gold Ref.}  & - & 0.9706 & 0.8790 & 0.7797 & 0.7077 & 0.8632 & 0.7754 \\
\bottomrule
\end{tabular}
}
\caption{Full evaluation results of our \textbf{Consistency} metrics on VWP narratives. \colorbox{mygrey2}{Notations are same as Table~\ref{tab:results_alignment_vwp_full}.}}
\label{tab:results_consistency_vwp_full}
\end{table}

\begin{table}[t]
\centering
\resizebox{0.75\columnwidth}{!}{
\smallskip\begin{tabular}{llcccc}
\toprule
\textbf{Model} & \textbf{Setting}  & \textbf{FID} & \textbf{CLIP-I} & \textbf{CLIP-T} \\
\toprule
\multirow{3}*{\textbf{ARLDM}} & No Constraint & 42.55 & 0.6384 & 0.1951 \\
                              & LLM Constraints & \textbf{37.60} & \textbf{0.6762} & \textbf{0.2036} \\
                              & Gold Constraints & \underline{35.25} & \underline{0.7156} & \underline{0.2089} \\
\midrule
\multirow{3}*{\textbf{StoryGen}} & No Constraint & 78.58 & 0.5624 & 0.1836 \\
                              & LLM Constraints & 52.09 & 0.6003 & 0.1935 \\
                              & Gold Constraints & 48.93 & 0.6194 & 0.1901 \\
\midrule
\multirow{6}*{\textbf{MM-Inter.}} & No Constraint & 48.33 & 0.6337 & 0.1758 \\
                              & LLM Constraints & 42.24 & 0.6670 & 0.1978 \\
                              & \quad - w/o CL & 42.85 & 0.6660 & 0.1966 \\
                              & \quad - w/o DS & 43.28 & 0.6568 & 0.1960 \\
                              & \quad - Random & 53.74 & 0.6143 & 0.1739 \\
                              & Gold Constraints & 39.27 & 0.6981 & 0.1997 \\
\midrule
\rowcolor{mygrey} \textbf{Gold Ref.}  & - & - & - & 0.2077 \\
\bottomrule
\end{tabular}
}
\caption{Evaluation results of full-reference metrics on VWP narratives. Lower FID is better. \colorbox{mygrey2}{Notations are same as Table~\ref{tab:results_alignment_vwp_full}.}}
\label{tab:results_other_vwp_full}
\end{table}

\begin{table*}[t]
\centering
\resizebox{1.0\textwidth}{!}{
\smallskip\begin{tabular}{l@{~~}lc@{~~}cc@{~~}cc@{~~}cc@{~~}cc@{~~}cc@{~~}c}
\toprule
\multirow{2}*{\textbf{Model}} & \multirow{2}*{\textbf{Setting}}  & \multicolumn{2}{c}{\textbf{Ranking}} & \multicolumn{2}{c}{\textbf{Non-Character}} & \multicolumn{2}{c}{\textbf{Character Number}} & \multicolumn{2}{c}{\textbf{Character Attribute}} & \multicolumn{2}{c}{\textbf{Time of Day}} & \multicolumn{2}{c}{\textbf{Location}}\\
    \cmidrule(lr){3-4}  \cmidrule(lr){5-6}  \cmidrule(lr){7-8}  \cmidrule(lr){9-10}  \cmidrule(lr){11-12}  \cmidrule(lr){13-14}
 &  & \textbf{CLIP-T-MRR} & \textbf{VQA-MRR} & \textbf{MiniCPM} & \textbf{Llava} & \textbf{MiniCPM} & \textbf{Llava} & \textbf{MiniCPM} & \textbf{Llava} & \textbf{MiniCPM} & \textbf{Llava} & \textbf{MiniCPM} & \textbf{Llava} \\
\toprule
\multirow{3}*{\textbf{ARLDM}} & No Constraint & 0.0954 & 0.1279 & 0.3487 & 0.3302 & 0.2682 & 0.2714 & 0.2330 & 0.2208 & 0.3250 & 0.2768 & 0.2987 & 0.3341 \\
                              & LLM Constraints & \textbf{0.1369} & \textbf{0.2273} & \textbf{0.6590} & \textbf{0.6233} & 0.4702 & \textbf{0.4330} & 0.3694 & 0.3434 & \textbf{0.4049} & 0.3623 & 0.4899 & 0.5031 \\
                              & Gold Constraints & \underline{0.1415} & \underline{0.2350} & \underline{0.6745} & \underline{0.6319} & 0.6067 & 0.5743 & 0.4804 & \underline{0.4485} & \underline{0.4689} & 0.4084 & 0.5994 & 0.6011 \\
\midrule
\multirow{3}*{\textbf{StoryGen}} & No Constraint & 0.0926 & 0.1079 & 0.3051 & 0.3080 & 0.2908 & 0.2956 & 0.2064 & 0.1593 & 0.1599 & 0.1988 & 0.1710 & 0.2505 \\
                              & LLM Constraints & 0.0992 & 0.1438 & 0.5259 & 0.5306 & 0.4304 & 0.3955 & 0.2684 & 0.2677 & 0.2754 & 0.2580 & 0.3739 & 0.4423 \\
                              & Gold Constraints & 0.1078 & 0.1653 & 0.5273 & 0.5291 & \underline{0.6629} & 0.5572 & 0.3709 & 0.3636 & 0.2950 & 0.2686 & 0.4281 & 0.4883 \\
\midrule
\multirow{3}*{\textbf{MM-Inter.}} & No Constraint & 0.0521 & 0.0979 & 0.3286 & 0.3264 & 0.2616 & 0.2323 & 0.2290 & 0.1956 & 0.3311 & 0.3042 & 0.2294 & 0.2588 \\
                              & LLM Constraints & 0.0983 & 0.1935 & 0.6030 & 0.5702 & \textbf{0.4767} & 0.4329 & \textbf{0.3796} & \textbf{0.3449} & 0.3733 & \textbf{0.3686} & \textbf{0.4971} & \textbf{0.5170} \\
                              & Gold Constraints & 0.1049 & 0.1959 & 0.6375 & 0.5786 & 0.6132 & \underline{0.5746} & \underline{0.4877} & 0.4456 & 0.4231 & \underline{0.4155} & \underline{0.6167} & \underline{0.6224} \\
\midrule
\rowcolor{mygrey} \textbf{Gold Ref.}  & - & 0.1657 & 0.2735 & 0.7630 & 0.7118 & 0.8682 & 0.8375 & 0.7981 & 0.7156 & 0.7620 & 0.7114 & 0.8955 & 0.7879 \\
\bottomrule
\end{tabular}
}
\caption{Full zero-shot evaluation results of our ranking-based and fine-grained \textbf{Alignment} metrics on Storyboard20K narratives. \colorbox{mygrey2}{Notations are same as Table~\ref{tab:results_alignment_vwp_full}.}}
\label{tab:results_alignment_sb20k_full}
\end{table*}

\begin{table}[t]
\centering
\resizebox{1.0\columnwidth}{!}{
\smallskip\begin{tabular}{@{~}l@{~~}lc@{~~}cc@{~~}cc@{~~}c@{~}}
\toprule
\multirow{2}*{\textbf{Model}} & \multirow{2}*{\textbf{Setting}}  & \multicolumn{2}{c}{\textbf{Style}} & \multicolumn{2}{c}{\textbf{Character}} & \multicolumn{2}{c}{\textbf{Location}}\\
    \cmidrule(lr){3-4}  \cmidrule(lr){5-6}  \cmidrule(lr){7-8}
 &  & \textbf{MiniCPM} & \textbf{Llava} & \textbf{MiniCPM} & \textbf{Llava} & \textbf{MiniCPM} & \textbf{Llava} \\
\toprule
\multirow{3}*{\textbf{ARLDM}} & No Constraint & 0.2279 & 0.2089 & 0.2459 & 0.2373 & 0.0879 & 0.1133 \\
                              & LLM Constraints & 0.6477 & 0.6140 & 0.5113 & 0.4531 & 0.3047 & 0.2686 \\
                              & Gold Constraints & 0.6968 & 0.6167 & 0.5997 & 0.5031 & 0.4219 & 0.3679 \\
\midrule
\multirow{3}*{\textbf{StoryGen}} & No Constraint & 0.2671 & 0.2613 & 0.0645 & 0.1033 & 0.2259 & 0.2108 \\
                              & LLM Constraints & 0.5209 & 0.5795 & 0.3156 & 0.3494 & 0.3526 & 0.3395 \\
                              & Gold Constraints & 0.5259 & 0.5848 & 0.3688 & 0.4108 & 0.4465 & 0.3975 \\
\midrule
\multirow{3}*{\textbf{MM-Inter.}} & No Constraint & 0.8766 & 0.8594 & 0.3627 & 0.3667 & 0.4207 & 0.3374 \\
                              & LLM Constraints & \textbf{0.9324} & \textbf{0.9016} & \textbf{0.6187} & \textbf{0.5694} & \textbf{0.6958} & \textbf{0.6378} \\
                              & Gold Constraints & \underline{0.9349} & \underline{0.9047} & \underline{0.6598} & \underline{0.6283} & \underline{0.7956} & \underline{0.7310} \\
\midrule
\rowcolor{mygrey} \textbf{Gold Ref.}  & - & 0.9399 & 0.8556 & 0.8118 & 0.7665 & 0.8955 &  0.7996\\
\bottomrule
\end{tabular}
}
\caption{Full zero-shot evaluation results of our \textbf{Consistency} metrics on Storyboard20K narratives. \colorbox{mygrey2}{Notations are same as Table~\ref{tab:results_alignment_vwp_full}.}}
\label{tab:results_consistency_sb20k_full}
\end{table}

\begin{table}[t]
\centering
\resizebox{0.75\columnwidth}{!}{
\smallskip\begin{tabular}{llcccc}
\toprule
\textbf{Model} & \textbf{Setting}  & \textbf{FID} & \textbf{CLIP-I} & \textbf{CLIP-T} \\
\toprule
\multirow{3}*{\textbf{ARLDM}} & No Constraint & 97.91 & 0.5910 & 0.1936 \\
                              & LLM Constraints & \textbf{82.64} & \textbf{0.6395} & \textbf{0.1995} \\
                              & Gold Constraints & \underline{77.70} & \underline{0.6754} & \underline{0.2057} \\
\midrule
\multirow{3}*{\textbf{StoryGen}} & No Constraint & 161.41 & 0.5367 & 0.1690 \\
                              & LLM Constraints & 112.03 & 0.5832 & 0.1880 \\
                              & Gold Constraints & 107.67 & 0.5966 & 0.1837 \\
\midrule
\multirow{3}*{\textbf{MM-Inter.}} & No Constraint & 102.42 & 0.5876 & 0.1644 \\
                              & LLM Constraints & 95.73 & 0.6362 & 0.1893 \\
                              & Gold Constraints & 90.82 & 0.6587 & 0.1933 \\
\midrule
\rowcolor{mygrey} \textbf{Gold Ref.}  & - & - & - & 0.2049 \\
\bottomrule
\end{tabular}
}
\caption{Evaluation results of full-reference metrics on Storyboard20K narratives. Lower FID is better. \colorbox{mygrey2}{Notations are same as Table~\ref{tab:results_alignment_vwp_full}.}}
\label{tab:results_other_sb20k_full}
\end{table}

\begin{table*}[t]
\centering
\resizebox{1.0\textwidth}{!}{
\smallskip\begin{tabular}{l@{~~}lc@{~~}cc@{~~}cc@{~~}cc@{~~}cc@{~~}cc@{~~}c}
\toprule
\multirow{2}*{\textbf{Model}} & \multirow{2}*{\textbf{Setting}}  & \multicolumn{2}{c}{\textbf{Ranking}} & \multicolumn{2}{c}{\textbf{Non-Character}} & \multicolumn{2}{c}{\textbf{Character Number}} & \multicolumn{2}{c}{\textbf{Character Attribute}} & \multicolumn{2}{c}{\textbf{Time of Day}} & \multicolumn{2}{c}{\textbf{Location}}\\
    \cmidrule(lr){3-4}  \cmidrule(lr){5-6}  \cmidrule(lr){7-8}  \cmidrule(lr){9-10}  \cmidrule(lr){11-12}  \cmidrule(lr){13-14}
 &  & \textbf{CLIP-T-MRR} & \textbf{VQA-MRR} & \textbf{MiniCPM} & \textbf{Llava} & \textbf{MiniCPM} & \textbf{Llava} & \textbf{MiniCPM} & \textbf{Llava} & \textbf{MiniCPM} & \textbf{Llava} & \textbf{MiniCPM} & \textbf{Llava} \\
\toprule
\multirow{3}*{\textbf{ARLDM}} & No Constraint & 0.1015 & 0.1367 & 0.4706 & 0.6048 & 0.2878 & 0.2884 & 0.1666 & 0.1764 & 0.4045 & 0.4135 & 0.3802 & 0.4041 \\
                              & LLM Constraints & 0.1428 & 0.2328 & \textbf{0.5685} & \textbf{0.6432} & 0.3065 & 0.3118 & 0.2217 & 0.2787 & 0.4409 & 0.4345 & 0.4420 & 0.4468 \\
                              & Gold Constraints & \underline{0.1493} & \underline{0.2417} & \underline{0.5771} & \underline{0.6519} & 0.3568 & 0.3386 & 0.2676 & 0.2984 & 0.4894 & 0.4474 & 0.4862 & 0.4839 \\
\midrule
\multirow{3}*{\textbf{StoryGen}} & No Constraint & 0.1010 & 0.1347 & 0.4536 & 0.5257 & 0.2825 & 0.2851 & 0.1651 & 0.1738 & 0.3965 & 0.4043 & 0.3735 & 0.3961 \\
                              & LLM Constraints & \textbf{0.1443} & \textbf{0.2348} & 0.5633 & 0.6037 & 0.3070 & 0.3148 & 0.2079 & 0.2559 & 0.4225 & 0.4267 & 0.3971 & 0.4205 \\
                              & Gold Constraints & 0.1469 & 0.2410 & 0.5714 & 0.6160 & 0.3515 & 0.3376 & 0.2577 & 0.2907 & 0.4690 & 0.4400 & 0.4385 & 0.4609 \\
\midrule
\multirow{3}*{\textbf{MM-Inter.}} & No Constraint & 0.0581 & 0.1062 & 0.4477 & 0.4983 & 0.2917 & 0.2946 & 0.1840 & 0.2188 & 0.4227 & 0.4158 & 0.3743 & 0.3714 \\
                              & LLM Constraints & 0.1065 & 0.2015 & 0.5352 & 0.5853 & \textbf{0.3662} & \textbf{0.3847} & \textbf{0.2645} & \textbf{0.2903} & \textbf{0.4727} & \textbf{0.4481} & \textbf{0.4587} & \textbf{0.4536} \\
                              & Gold Constraints & 0.1124 & 0.2032 & 0.5450 & 0.5986 & \underline{0.4126} &  \underline{0.4242} & \underline{0.3125} & \underline{0.3238} & \underline{0.5030} & \underline{0.4624} & \underline{0.5609} & \underline{0.5375} \\
\midrule
\rowcolor{mygrey} \textbf{Gold Ref.}  & - & 0.1601 & 0.2688 & 0.7584 & 0.7432 & 0.8171 & 0.8061 & 0.7780 & 0.7655 & 0.7545 & 0.7728 & 0.7523 & 0.7635 \\
\bottomrule
\end{tabular}
}
\caption{Full evaluation results of our ranking-based and fine-grained \textbf{Alignment} metrics on StorySalon narratives. \colorbox{mygrey2}{Notations are same as Table~\ref{tab:results_alignment_vwp_full}.}}
\label{tab:results_alignment_ssalon_full}
\end{table*}

\begin{table}[t]
\centering
\resizebox{1.0\columnwidth}{!}{
\smallskip\begin{tabular}{@{~}l@{~~}lc@{~~}cc@{~~}cc@{~~}c@{~}}
\toprule
\multirow{2}*{\textbf{Model}} & \multirow{2}*{\textbf{Setting}}  & \multicolumn{2}{c}{\textbf{Style}} & \multicolumn{2}{c}{\textbf{Character}} & \multicolumn{2}{c}{\textbf{Location}}\\
    \cmidrule(lr){3-4}  \cmidrule(lr){5-6}  \cmidrule(lr){7-8}
 &  & \textbf{MiniCPM} & \textbf{Llava} & \textbf{MiniCPM} & \textbf{Llava} & \textbf{MiniCPM} & \textbf{Llava} \\
\toprule
\multirow{3}*{\textbf{ARLDM}} & No Constraint & 0.5000 & 0.4824 & 0.1461 & 0.1839 & 0.2903 & 0.2596 \\
                              & LLM Constraints & 0.6563 & 0.5684 & 0.2622 & 0.2551 & 0.3296 & 0.2978 \\
                              & Gold Constraints & 0.6875 & 0.5770 & 0.2890 & 0.2672 & 0.3839 & 0.3257 \\
\midrule
\multirow{3}*{\textbf{StoryGen}} & No Constraint & 0.4246 & 0.4197 & 0.1041 & 0.1362 & 0.2265 & 0.2205 \\
                              & LLM Constraints & 0.6073 & 0.5583 & 0.2886 & 0.2793 & 0.3191 & 0.2784 \\
                              & Gold Constraints & 0.6472 & 0.5609 & 0.2911 & 0.2826 & 0.3745 & 0.3147 \\
\midrule
\multirow{3}*{\textbf{MM-Inter.}} & No Constraint & 0.9450 & 0.8668 & 0.3349 & 0.4086 & 0.7022 & 0.6232 \\
                              & LLM Constraints & \textbf{0.9563} & \textbf{0.8747} & \textbf{0.3545} & \textbf{0.4449} & \textbf{0.7798} & \textbf{0.6978} \\
                              & Gold Constraints & \underline{0.9688} & \underline{0.8786} & \underline{0.3834} & \underline{0.4737} & \underline{0.8034} & \underline{0.7617} \\
\midrule
\rowcolor{mygrey} \textbf{Gold Ref.}  & - & 0.9688 & 0.9865 & 0.7686 & 0.7611 & 0.8135 & 0.8059 \\
\bottomrule
\end{tabular}
}
\caption{Full evaluation results of our \textbf{Consistency} metrics on StorySalon narratives. \colorbox{mygrey2}{Notations are same as Table~\ref{tab:results_alignment_vwp_full}.}}
\label{tab:results_consistency_ssalon_full}
\end{table}

\begin{table}[t]
\centering
\resizebox{0.75\columnwidth}{!}{
\smallskip\begin{tabular}{llcccc}
\toprule
\textbf{Model} & \textbf{Setting}  & \textbf{FID} & \textbf{CLIP-I} & \textbf{CLIP-T} \\
\toprule
\multirow{3}*{\textbf{ARLDM}} & No Constraint & 64.69 & 0.6278 & 0.1975 \\
                              & LLM Constraints & 56.65 & 0.6515 & 0.2001 \\
                              & Gold Constraints & 56.51 & 0.6887 & \underline{0.2022} \\
\midrule
\multirow{3}*{\textbf{StoryGen}} & No Constraint & 63.63 & 0.6463 & 0.1946 \\
                              & LLM Constraints & \textbf{56.18} & \textbf{0.6600} & \textbf{0.2005} \\
                              & Gold Constraints & \underline{55.62} & \underline{0.6919} & 0.2021 \\
\midrule
\multirow{3}*{\textbf{MM-Inter.}} & No Constraint & 74.92 & 0.6370 & 0.1834 \\
                              & LLM Constraints & 72.91 & 0.6552 & 0.1879 \\
                              & Gold Constraints & 72.03 & 0.6780 & 0.1896 \\
\midrule
\rowcolor{mygrey} \textbf{Gold Ref.}  & - & - & - & 0.2065 \\
\bottomrule
\end{tabular}
}
\caption{Evaluation results of full-reference metrics on StorySalon narratives. Lower FID is better. \colorbox{mygrey2}{Notations are same as Table~\ref{tab:results_alignment_vwp_full}.}}
\label{tab:results_other_ssalon_full}
\end{table}

We test three representative visual narrative generation models on \ourbench{}, which cover diverse model structures, as described below:
\begin{itemize}
\item \textbf{ARLDM} \citep{pan2024synthesizing} trains a Stable Diffusion \citep{rombach2022high} module to auto-regressively generate each visual narrative image, which is conditioned on the BLIP \citep{li2022blip} embeddings of previous scenes' generated images and input textual constraints, and the CLIP \citep{radford2021learning} embedding of current scene's input textual constraints.
\item \textbf{StoryGen} \citep{liu2024intelligent} uses a dual-diffusion structure to perform the auto-regressive generation of narrative images.
It first adds noise to each previously generated image, and then the noisy image is de-noised by a Stable Diffusion module (conditioned on the image's corresponding input textual constraints), whose latent diffusion states are used as the extracted features of the image.
Conditioned on the current textual constraints and the concatenation of previous images' extracted features, a second Stable Diffusion module is trained to generate the current narrative image.
\item \textbf{MM-Interleaved (MM-Inter.)} \citep{tian2024mm} trains a VLM, \ie{}, Vicuna \citep{zheng2023judging} with CLIP vision encoder, to model the interleaved sequence of previously generated images and their textual constraints, and a Stable Diffusion module to generate the current narrative image based on the output states of the VLM.
Both the VLM and the diffusion module are augmented by additional layers of cross-attention to sparse image features via Deformable Attention \citep{zhu2020deformable}.
\end{itemize}

%
%
%

\section{Full Experimental Results}
\label{sec:full_results}

Table~\ref{tab:results_alignment_vwp_full}~-~\ref{tab:results_other_ssalon_full} present the full evaluation results of visual narrative generation on \ourbench{}.
All results coherently indicate the same conclusion that learning with \ourbench{}'s commonsense and discourse constraints significantly improves the consistency of visual narrative generations and their alignment to the input textual narrative.
Moreover, our two ranking-based metrics CLIP-T-MRR and VQA-MRR consistently show that all model generations and the gold reference score far below the maximum (1.0), supporting the fact that creating visual narratives is a considerably open-ended task, which does not possess the only feasible reference that always ranks the first.
More importantly, our VQA-based metrics deployed on MiniCPM-V-2.6 and LLaVA-OneVision-72B demonstrate mostly aligned preference among different models and settings.
This verifies that our proposed metrics are not biased on the preference of a specific VLM used for generating the VQA scores.

Besides of MM-Interleaved, which is the best-performed model fine-tuned on \ourbench{}, we further test other similar interleaved image-text generative models, including \textbf{Anole} \citep{chern2024anole} and \textbf{Lumina-mGPT} \citep{liu2024lumina}, which however completely fail our benchmark task (with nearly zero scores on \ourbench{} metrics) under zero-shot or few-shot settings.\footnote{We verify that MM-Interleaved model would also fail our benchmark task under zero/few-shot settings, \ie{}, without fine-tuning.}
This indicates that supervised learning (or fine-tuning) is necessary for current interleaved image-text generative models to address our benchmark's challenging task, while the fine-tuning codes of these models are not publicly available, which hinders more experimental verifications.

\begin{figure}[t]
\centering
\includegraphics[width=1.0\columnwidth]{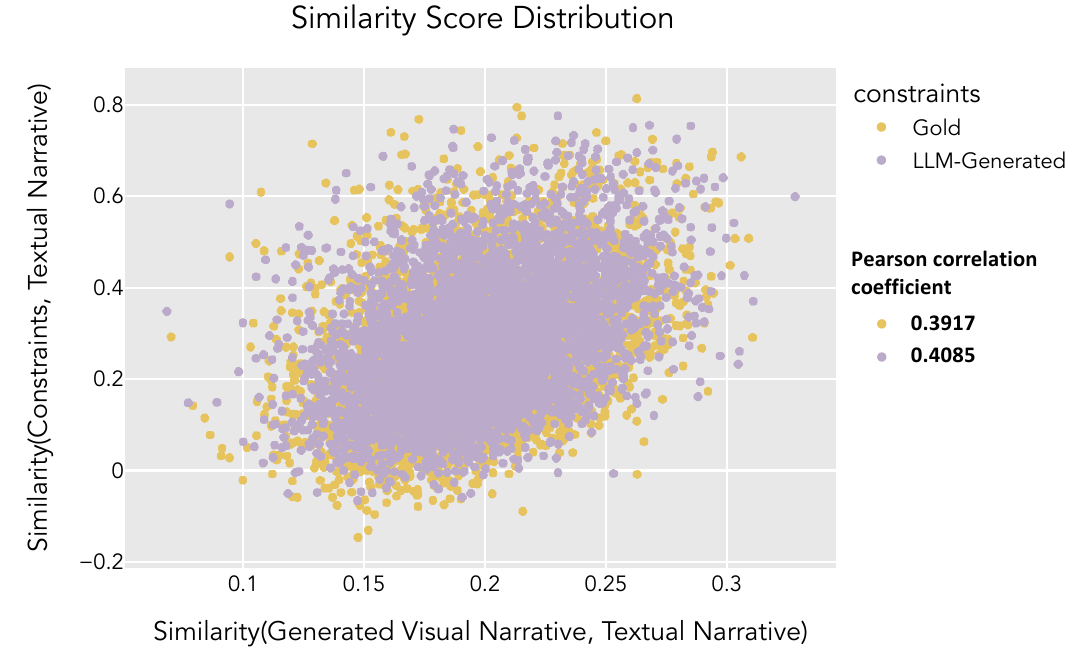}
\caption{Correlation between generated visual narrative images and augmented narrative constraints (either from gold labels or generated by LLM, Llama3.1-70B-Instruct), \wrt{} their CLIP embedding similarity to the input textual narrative. Data samples are from MM-Interleaved generations (with LLM Constraints and with Gold Constraints) on VWP narratives.}
\label{alignment_correlation}
\end{figure}

Figure~\ref{alignment_correlation} shows the distribution of paired similarity scores in our correlation study between visual narrative generation and constraints, where the x-axis denotes the CLIP similarity between each visual generation and input textual narrative, and the y-axis denotes the CLIP similarity between the sample's augmented constraints and the textual narrative.
The distribution demonstrates a clear positive correlation between the narrative constraints and their resulting visual narrative generations, with $\sim 0.4$ Pearson correlation coefficient, no matter whether the constraints are from gold labels or generated by LLM.
This highlights the importance of planning faithful storytelling constraints to advance visual narrative generations.

We also evaluate MM-Interleaved model on varied settings of using LLMs to generate narrative constraints (with LLM Constraints), including 4-shot (\textbf{4S}) prompting Llama3.1-70B-Instruct (\textbf{Llama-70B}), and fine-tuning (\textbf{FT}) Llama3.1-8B-Instruct (\textbf{Llama-8B}), \textbf{Gemma-7B} and \textbf{Qwen2-7B}, compared to our adopted setting of fine-tuning Llama3.1-70B-Instruct with LoRA.
Results in Table~\ref{tab:ablation_llm_cons}, based on the VWP narratives of \ourbench{}, show that our adopted setting best augments visual narrative generation.

\begin{table}[t]
\centering
\resizebox{1.0\columnwidth}{!}{
\smallskip\begin{tabular}{@{~}l@{~~~~}c@{~~~~}c@{~~~~}c@{~~~~}c@{~~~~}c@{~~~~}c@{~}}
\toprule
\textbf{LLM Constraints}  & \textbf{FID} & \textbf{CLIP-I} & \textbf{CLIP-T} & \textbf{CLIP-T-MRR} & \textbf{Alignment} & \textbf{Consistency}\\
\toprule
\textbf{FT Llama-70B} & 42.24 & 0.6670 & 0.1978 & 0.1107 & 0.5197 & 0.8093 \\
\textbf{4S Llama-70B} & 42.95 & 0.6625 & 0.1973 & 0.1104 & 0.4948 & 0.7936 \\
\textbf{FT Llama-8B} & 49.61 & 0.6293 & 0.1833 & 0.0570 & 0.3980 & 0.7436 \\
\textbf{FT Gemma-7B} & 51.69 & 0.6180 & 0.1788 & 0.0445 & 0.3751 & 0.7312 \\
\textbf{FT Qwen2-7B} & 47.83 & 0.6376 & 0.1915 & 0.0866 & 0.4507 & 0.7606 \\
\midrule
\rowcolor{mygrey} \textbf{Gold Ref.}  & -  & - & 0.2077 & 0.1586 & 0.7930 & 0.8711 \\
\bottomrule
\end{tabular}
}
\caption{Performance of MM-Interleaved model with different LLM-generated narrative constraints, evaluated on VWP narratives. Llama3.1-70B-Instruct (Llama-70B) is fine-tuned (FT) with LoRA or 4-shot (4S) prompted, while Llama3.1-8B-Instruct (Llama-8B), Gemma-7B and Qwen2-7B are fully fine-tuned. \textbf{Alignment} and \textbf{Consistency} denote the average score of our proposed fine-grained alignment and consistency metrics.
}
\label{tab:ablation_llm_cons}
\end{table}

\begin{figure*}[t]
\centering
\includegraphics[width=1.0\textwidth]{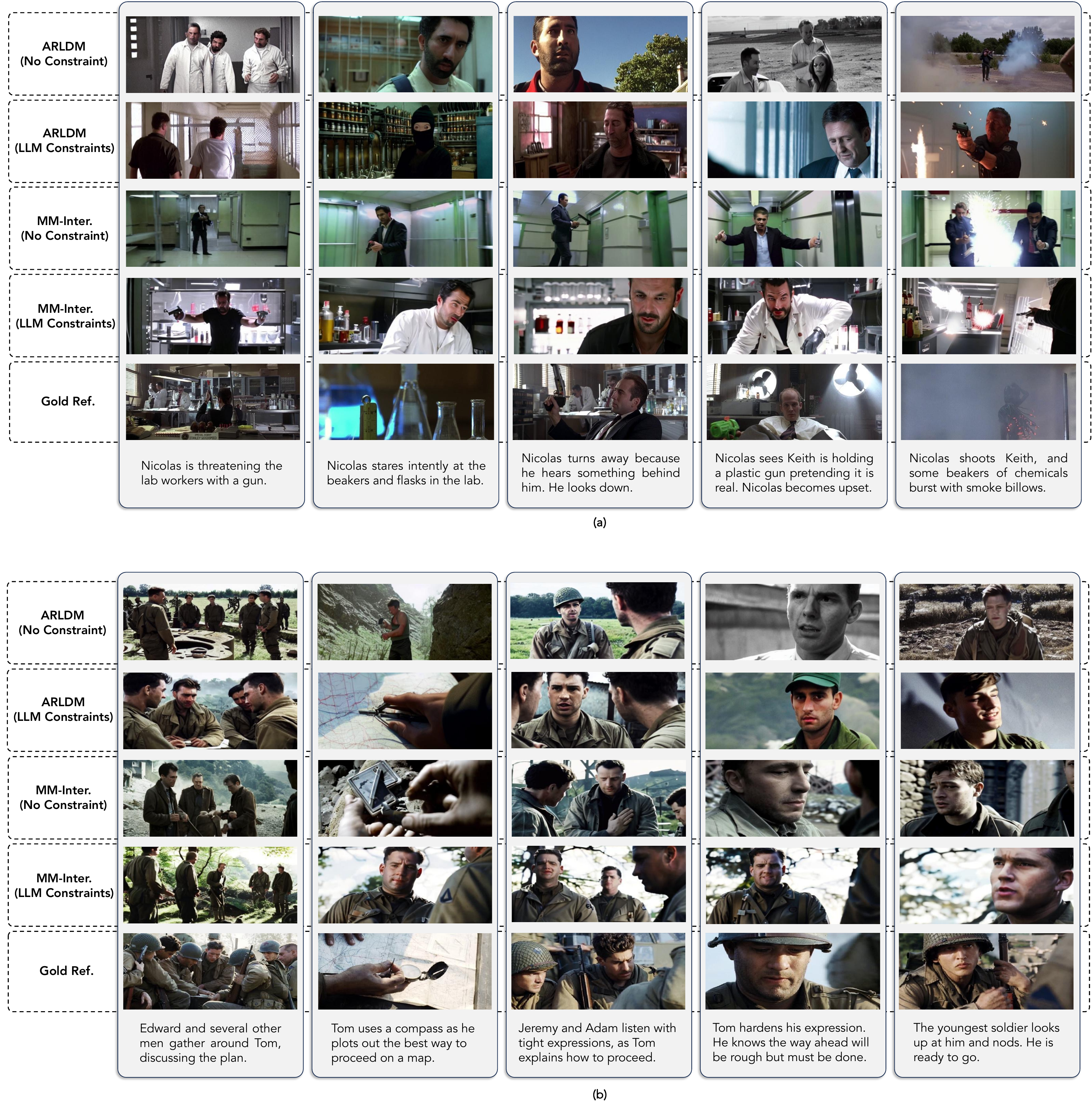}
\caption{Visual narratives generated by ARLDM and MM-Interleaved (MM-Inter.), with and without LLM-generated narrative constraints, compared to the gold reference.
In narrative (a), LLM-generated constraints significantly improve MM-Interleaved, by pushing its generation more aligned with the lab setting described in textual narrative.
By contrast, ARLDM fails to generate images with decent alignment to textual narrative, although the image style consistency is improved by LLM constraints, \eg{}, avoid generating a black and white image at the fourth scene.
In narrative (b), the generation of ARLDM with LLM constraints turns out to achieve improved image style consistency and alignment to textual narrative plot, \eg{}, showing a map in the second scene.
Besides, compared to MM-Interleaved without constraint, the generation of MM-Interleaved with LLM constraints displays better consistency of character (\eg{}, Tom) facial features and background location, and comparable faithfulness to textual narrative.
However, both model generations with constraints still contain unreasonable contents, \eg{}, a sudden shift of character Nicolas's outfit in the generation of MM-Interleaved (with LLM Constraints) in (a), inconsistent faces of character Tom in the ARLDM (with LLM Constraints) generation in (b).
}
\label{full_case_study}
\end{figure*}

Figure~\ref{full_case_study} displays several visual narratives generated by our deployed baseline methods.
The model generations still contain unfaithful or inconsistent contents, even with the augmentation of narrative constraints.
This reveals the challenge of developing more robust methods for the visual narrative generation, which we leave for future work.

\begin{figure*}[t]
\begin{subfigure}{1.0\textwidth}
  \centering
  \includegraphics[width=0.8\linewidth]{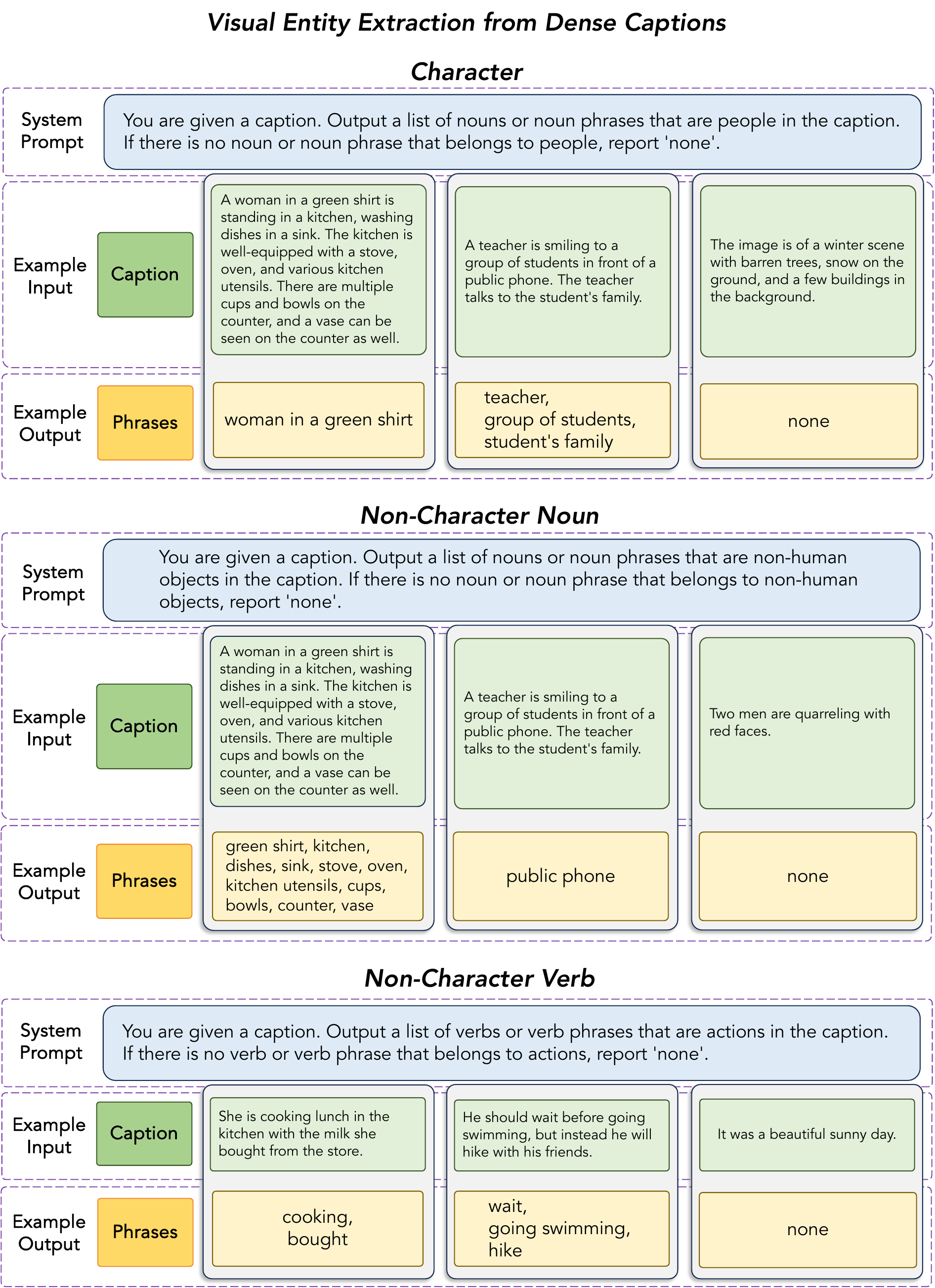}
  \caption{}
  \label{commonsense_linking_1}
\end{subfigure}
\end{figure*}
\begin{figure*}[t] \ContinuedFloat 
\begin{subfigure}{1.0\textwidth}
  \centering
  \includegraphics[width=0.8\linewidth]{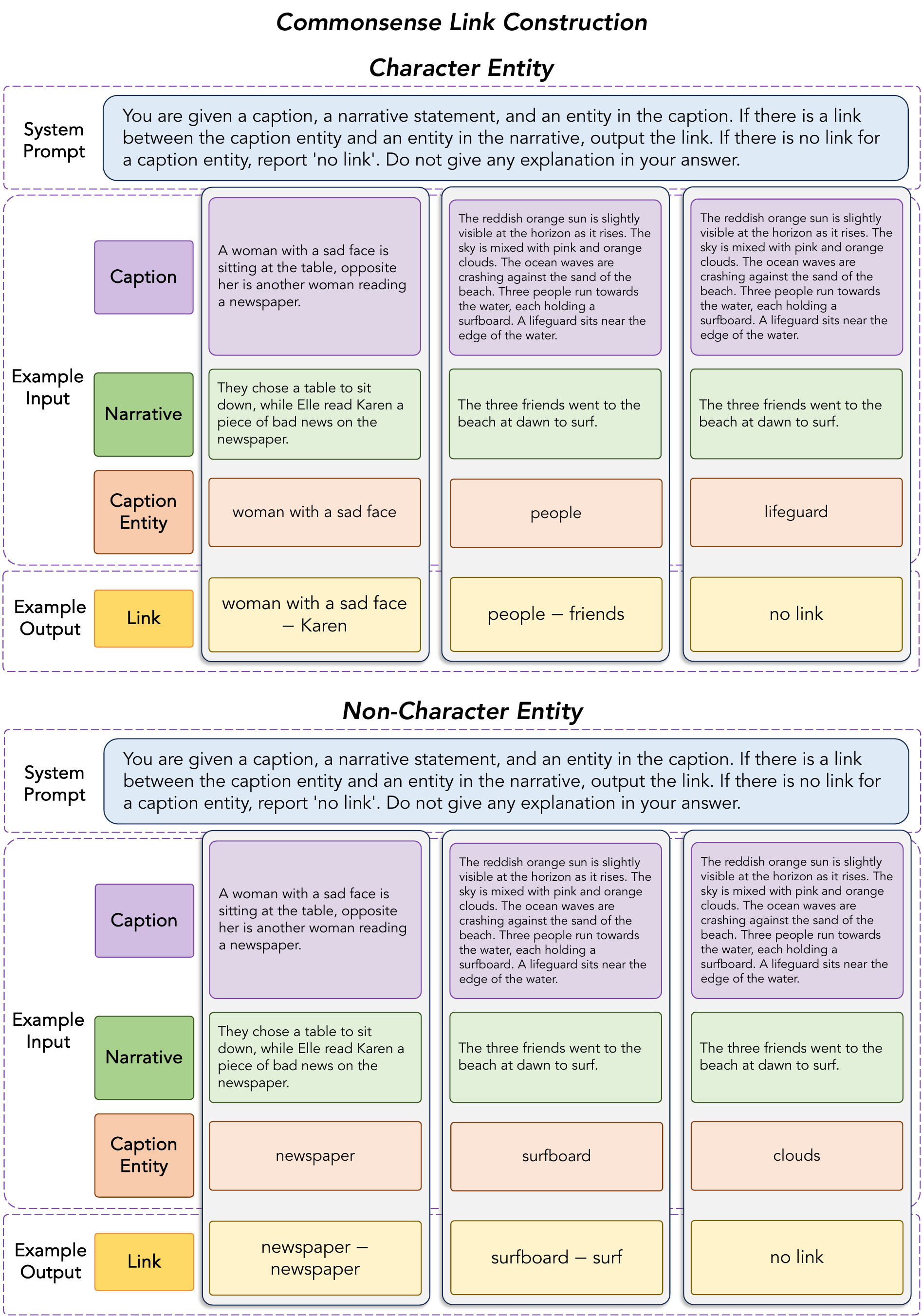}
  \caption{}
  \label{commonsense_linking_2}
\end{subfigure}
\caption{Few-shot prompting demonstrations for constructing the \textbf{commonsense links} in \ourbench{}, including (a) visual entity extraction (\wrt{} character, non-character noun and verb), and (b) link construction (\wrt{} each extracted character and non-character entity).}
\label{commonsense_linking}
\end{figure*}




\begin{figure*}[t]
\begin{subfigure}{1.0\textwidth}
  \centering
  \includegraphics[width=0.8\linewidth]{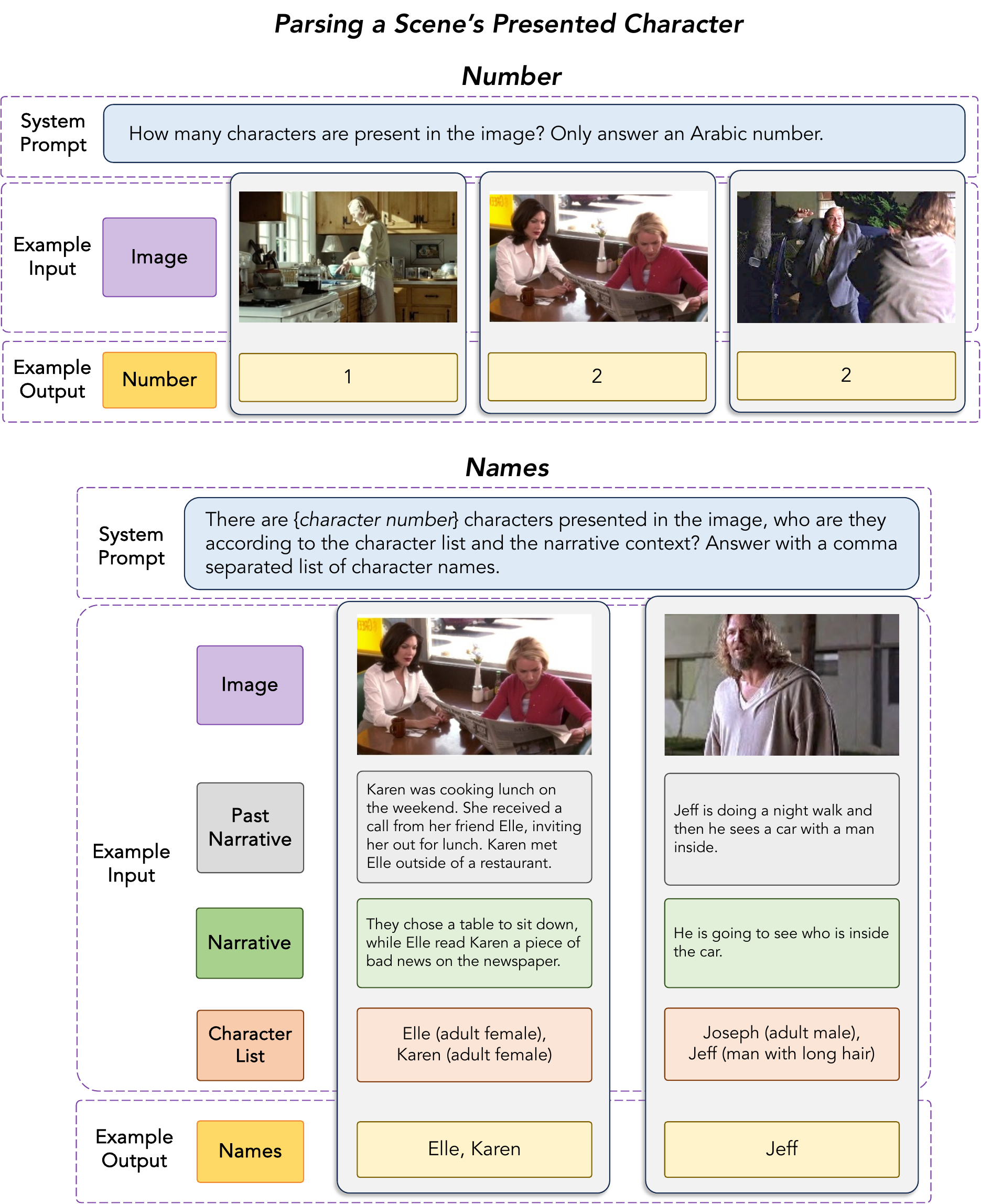}
  \caption{}
  \label{scene_features_1}
\end{subfigure}
\end{figure*}
\begin{figure*}[t] \ContinuedFloat 
\begin{subfigure}{1.0\textwidth}
  \centering
  \includegraphics[width=0.84\linewidth]{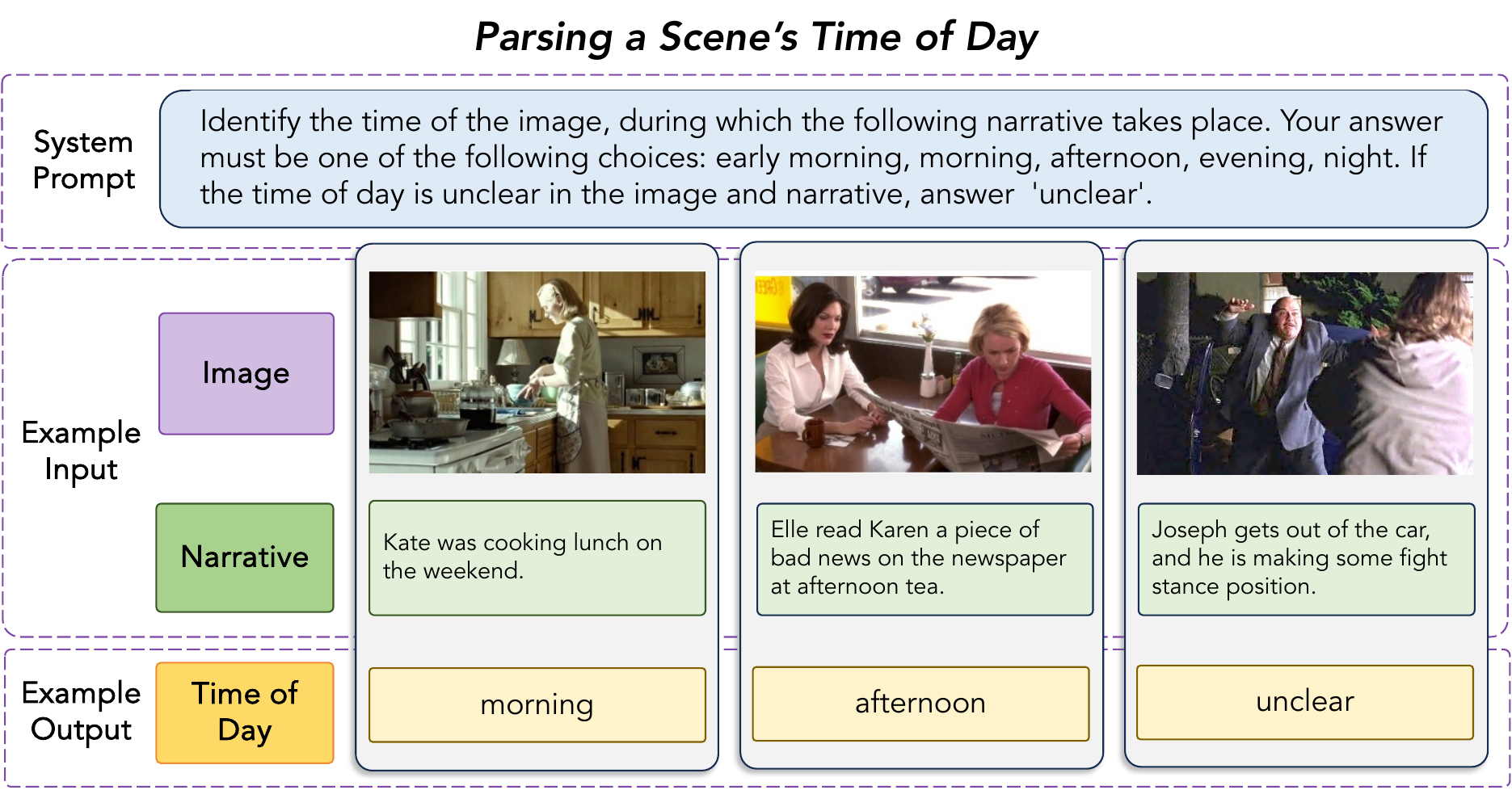}
  \caption{}
  \label{scene_features_2}
\end{subfigure}
\begin{subfigure}{1.0\textwidth}
  \centering
  \includegraphics[width=0.58\linewidth]{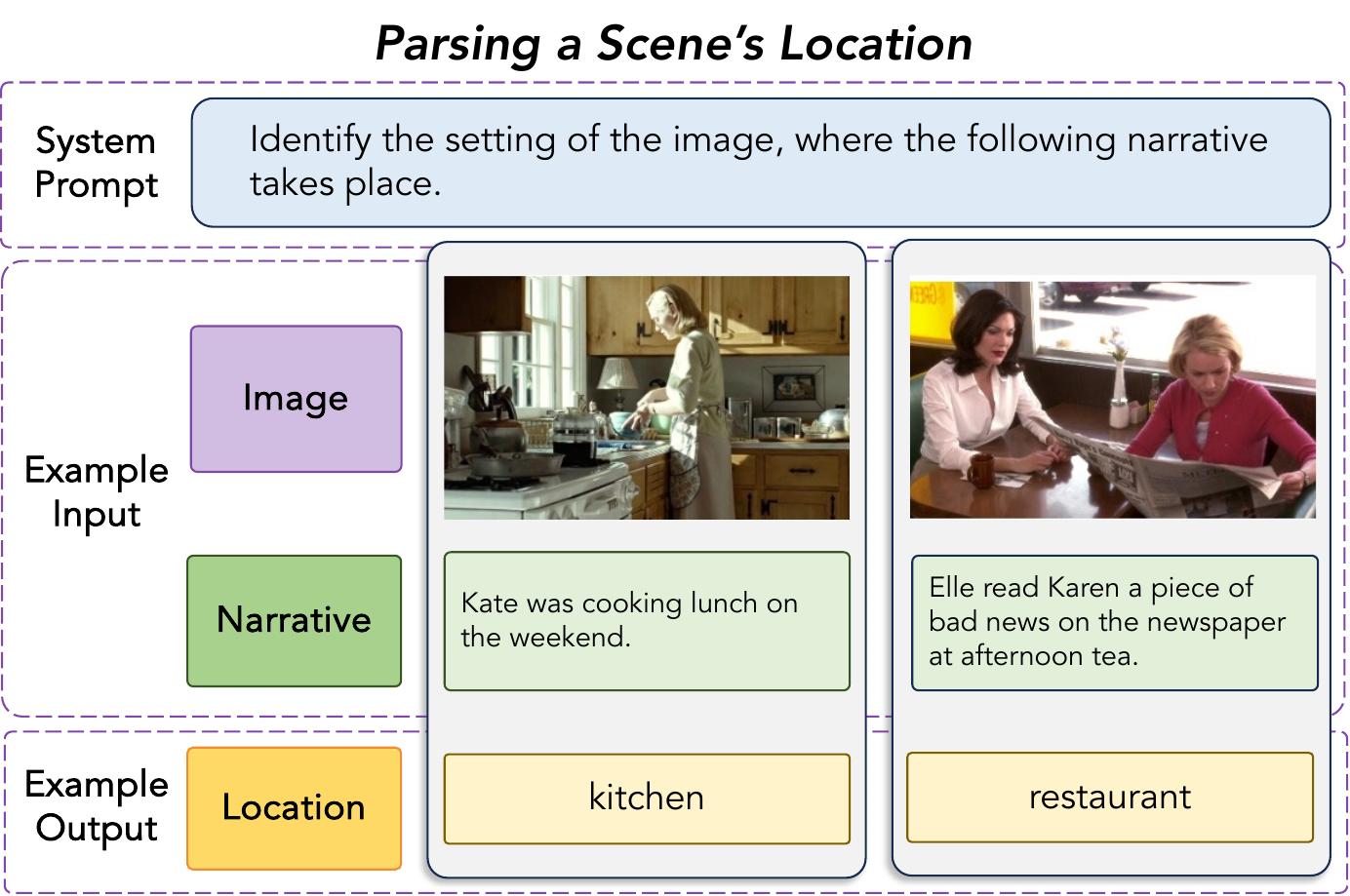}
  \caption{}
  \label{scene_features_3}
\end{subfigure}
\caption{Few-shot prompting demonstrations for parsing the \textbf{scene features} in \ourbench{}, including (a) presented character number and names, (b) time of day, and (c) location. In the step of parsing presented character names in (a), the span ``\{\textit{character number}\}'' in the system prompt is replaced by the output in the prior step of parsing character number.}
\label{scene_features}
\end{figure*}


\end{document}